\documentclass[10pt,journal,compsoc]{IEEEtran}
\ifCLASSOPTIONcompsoc
  \usepackage[nocompress]{cite}
\else
  \usepackage{cite}
\fi
\ifCLASSINFOpdf
  \usepackage[pdftex]{graphicx}
\else
\fi
\usepackage{amsmath}
\DeclareMathOperator*{\argmax}{argmax}

\usepackage{algorithmic}

\usepackage{array}

\ifCLASSOPTIONcompsoc
  \usepackage[caption=false,font=footnotesize,labelfont=sf,textfont=sf]{subfig}
\else
  \usepackage[caption=false,font=footnotesize]{subfig}
\fi
\usepackage{url}

\usepackage{multirow}
\usepackage{xcolor}
\usepackage{tikz}
\usepackage{pgfplots}
\usepackage{amssymb}
\usepackage{booktabs}

\renewcommand\paragraph[1]{\textbf{#1.}}

\usepackage{tikzpagenodes}

\begin{document}
%
\title{The  EuroCity Persons Dataset:\newline A Novel Benchmark for Object Detection}

\author{Markus Braun, Sebastian Krebs, Fabian Flohr, and Dariu M. Gavrila
\IEEEcompsocitemizethanks{\IEEEcompsocthanksitem M. Braun, S. Krebs and F. Flohr are with the Environment Perception Group, Daimler AG 
\IEEEcompsocthanksitem M. Braun, S. Krebs, and D. M. Gavrila are with the Intelligent Vehicles Group at TU Delft.}
}

\input{bib.short.def}


\IEEEtitleabstractindextext{%

\begin{abstract}
Big data has had a great share in the success of deep learning in computer vision. Recent works suggest that there is significant further potential to increase object detection performance by utilizing even bigger datasets. In this paper, we introduce the EuroCity Persons dataset, which provides a large number of highly diverse, accurate and detailed annotations of pedestrians, cyclists and other riders in urban traffic scenes. The images for this dataset were collected on-board a moving vehicle in 31 cities of 12 European countries. With over 238200 person instances manually labeled in over 47300 images, EuroCity Persons is nearly one \emph{order} of magnitude larger than person datasets used previously for benchmarking. The dataset furthermore contains a large number of person orientation annotations (over 211200).
We optimize four state-of-the-art deep learning approaches (Faster R-CNN, R-FCN, SSD and YOLOv3) to serve as baselines for the new object detection benchmark. In experiments with previous datasets we analyze the generalization capabilities of these detectors when trained with the new dataset. We furthermore study the effect of the training set size, the dataset diversity (day- vs. night-time, geographical region), the dataset detail (i.e. availability of object orientation information) and the annotation quality on the detector performance. Finally, we analyze error sources and discuss the road ahead.
\end{abstract}

}

\maketitle

\IEEEdisplaynontitleabstractindextext

%
\IEEEpeerreviewmaketitle


%
%
%
%

 

\IEEEraisesectionheading{\section{Introduction}\label{sec:introduction}}

\IEEEPARstart{P}{erson} detection in images is a key task in a number of important application domains, such as intelligent vehicles, surveillance, and robotics. Despite two decades of steady progress, it is still an open research problem. The wide variation in person appearance, arising from articulated pose, clothing, background and visibility conditions (time of day, weather), makes person detection particularly challenging. It therefore often features as canonical task to assess the performance of generic object detectors.

In this paper, we focus on the application setting of detecting persons in urban traffic scenes, as observed from cameras on-board a moving vehicle.  Detection performance has improved to the point that pedestrian and cyclist detection is incorporated in active safety systems of various premium vehicles on the market. Still, such systems are deployed in the context of driver assistance, meaning that a correct detection performance of about 90\% is acceptable, as long as the false alarm rate is essentially zero. With the advent of fully self-driving vehicles, performance needs to be significantly upped, as a driver is no longer available. A recent paper \cite{zhang2016tpami} argues that current pedestrian detection performance lags that of an attentive human by an order of magnitude. How can this performance gap be closed?

\begin{figure}[t]
\begin{center}
   \includegraphics[width=0.95\linewidth]{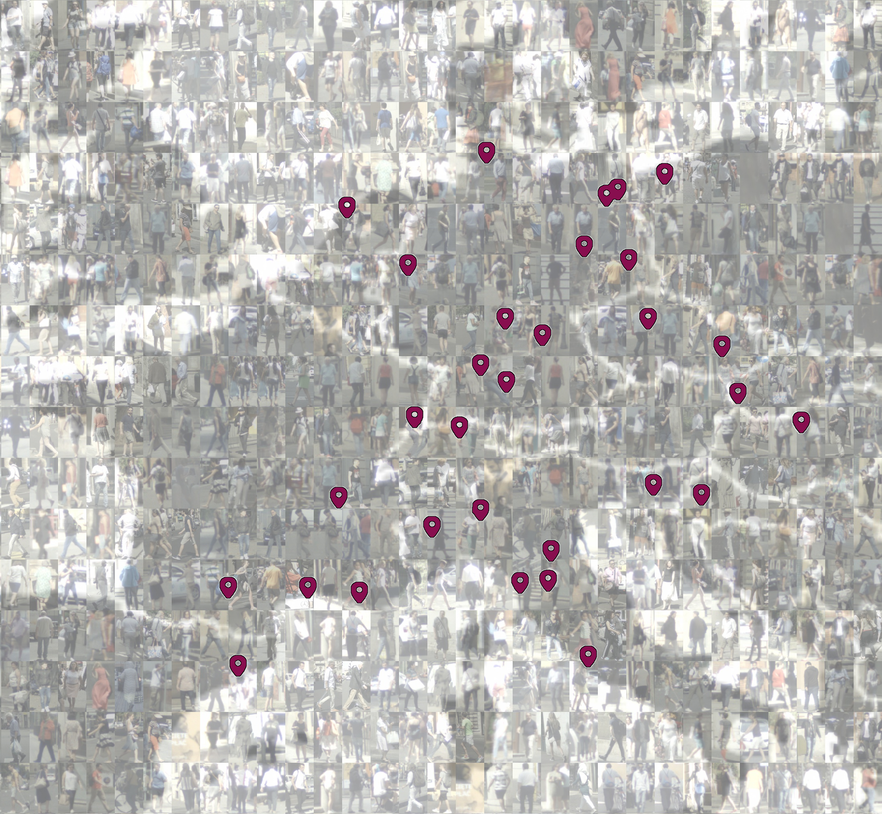}
\end{center}
   \caption{The EuroCity Persons dataset was recorded in 31 cities of 12 European countries: Croatia (Zagreb), Czech Republic (Brno, Prague), France (Lyon, Marseille, Montpellier, Toulouse), Germany (Berlin, Dresden, Hamburg, K\"oln, Leipzig, N\"urnberg, Potsdam, Stuttgart, Ulm and W\"urzburg), Hungary (Budapest), Italy (Bologna, Firenze, Milano, Pisa, Roma and Torino), The Netherlands (Amsterdam), Poland (Szczecin), Slovak Republic (Bratislava), Slovania (Ljubljana), Spain (Barcelona) and Switzerland (Basel, Z\"urich). The map itself was compiled from 500 randomly sampled pedestrian bounding boxes from our dataset.}
	\label{fig:EuroCity-map}
\end{figure}

\begin{table*}[!htp]
\small
\vspace*{5pt}
\caption{Comparison of person detection benchmarks in vehicle context} 
\label{tab:benchmarks_det}
\centering
\begin{tabular}{@{} l c c c c c @{}}
\toprule
& Caltech \cite{dollar2012pedestrian} & KITTI \cite{Geiger2012} & CityPersons \cite{ben2017CityPerson} & TDC \cite{Li2016} & \textbf{EuroCity Persons} \\
\hline
\# countries       		&  1 & 1 & 3  & 1 & \textbf{12} \\
\# cities          		&  1 & 1 & 27 & 1 & \textbf{31} \\
\# seasons        		&  1 & 1 & 3  & 1 & \textbf{4}  \\
\# images (day / night) &  \textbf{249884} / -  & 14999 / - & 5000 / - & 14674 / - & 40219 / \textbf{7118} \\
\# pedestrians (day / night)  & \textbf{289395}\footnotemark[1] / - & $\sim$9400\footnotemark[2] / - & 31514 / - & 8919 / - & 183182 / \textbf{35323}               \\ 
\# riders (day / night)     & - / - & $\sim$3300\footnotemark[2] / - & 3502 / - & \textbf{23442} / -  & 18216 / \textbf{1564} \\
\# ignore regions (day / night) & 57226\footnotemark[1] / - & $\sim$22600\footnotemark[2] / - & 13172 / - & - / - & \textbf{75495} / \textbf{20018} \\
\# orientations (day / night) & - / - & $\sim$12700\footnotemark[2] / - & - / - & - / - & \textbf{176901} / \textbf{34394} \\ 
resolution       		& $640\times480$ & $1240\times376$ & \textbf{2048}$\times$\textbf{1024} & \textbf{2048}$\times$\textbf{1024} & $1920\times1024$ \\
weather 				& dry & dry & dry & dry & \textbf{dry, wet} \\													
train-val-test split (\%) 	& 50-0-50 & 50-0-50 & 60-10-30 & 71-8-21 & 60-10-30 \\
\bottomrule
\end{tabular}
\end{table*}

Datasets play a crucial role in today's computer vision research \cite{deng2009imagenet}. Corresponding benchmarks reveal strengths and weaknesses of existing approaches and are instrumental in guiding research forward. 
Still, \cite{sun2017} argues that even larger datasets are needed. Experiments on their 300 million images dataset show that the classification performance further increases logarithmically with the size of the training dataset. Deep learning has also been very successful in the context of object detection \cite{Girshick2014} \cite{Girshick2015} and \cite{Ren2015}. 
More data could prove useful for object detection as well \cite{ben2017CityPerson}.
During the last two decades an extensive amount of research has been spent on pedestrian detection \cite{enzweiler2009monocular},\cite{dollar2012pedestrian},\cite{benenson2014ten},\cite{Hosang2015DnnForPedestrians},\cite{zhang2016tpami},\cite{rajaram2015exploration}. For several years, progress in this domain was monitored on  benchmarks like Caltech \cite{dollar2012pedestrian} and KITTI \cite{Geiger2012}. However, these datasets have come into age since. The recording conditions back then (i.e. image resolution and quality) do not reflect the current state of the art anymore. The comparatively small size of the training data (i.e. several thousands samples) furthermore makes these benchmarks prone to dataset bias and to over-fitting \cite{torralba2011}. Recently, CityPersons \cite{ben2017CityPerson} was released with higher resolution images and a larger quantity of training data ($\approx 35000$ person samples). Although these data additions are helpful, \cite{ben2017CityPerson} conclude that more training data is necessary for the recent high-capacity deep learning architectures. Data diversity is another important aspect. The before-mentioned datasets were captured in few countries ($1-3$), and in daylight and dry weather conditions only; this hampers generalization to real world applications.

\footnotetext[1]{Only an unspecified subset of these annotations were done manually, the remainder was obtained by interpolation (we estimate the number of manual annotations to be an order of magnitude smaller).}
\footnotetext[2]{Number estimated on the basis of the average number of pedestrians per image, since the test set is private and the authors did not report the actual number.}

To address these limitations we introduce a new dataset for vision-based person detection coined EuroCity Persons. The images for this dataset were collected on-board a moving vehicle in 31 cities of 12 European countries, see Figure \ref{fig:EuroCity-map} and Table \ref{tab:qualitative_true_positives}. With over 238200 person instances manually labeled in over 47300 images, EuroCity Persons is nearly one order of magnitude larger than person datasets used previously for benchmarking, in terms of manual annotations. Due to its comparatively large geographic coverage, its recordings during both day and night-time, and during all four seasons (light/short summer to thick/long winter clothing) it provides a new level of data diversity. EuroCity Persons furthermore offers detailed annotations; besides bounding box information, it includes tags for occlusion/truncation and annotates body orientation (the latter has relevance for object tracking and path prediction). Finally, thanks to the implemented quality control procedures, annotations are overall accurate.

By means of an experimental study using EuroCity Persons, we address a number of questions: how much do recent deep learning methods improve by an increased amount of training data? How well does this dataset generalize to existing datasets? What is the day- and night-time performance? Is there a geographical bias? How does annotation quality affect object detection performance? Does multi-tasking (orientation estimation) help object detection?

\section{Related Work}
\label{sec:related}

\subsection{Datasets}

A number of early datasets focus on pedestrian classification (e.g. Daimler-CB \cite{munder2006experimental}, CVC \cite{geronimo2007adaptive}, and NICTA \cite{overett2008new}) and detection (e.g. Daimler-DB \cite{enzweiler2009monocular}, INRIA \cite{dalal2005histograms}, ETH \cite{ess2007depth}, and TUD-Brussels \cite{wojek2009multi}). See \cite{enzweiler2009monocular} for an overview. Currently, KITTI \cite{Geiger2012} and the Caltech \cite{dollar2012pedestrian} are the established pedestrian detection benchmarks. The latter has been extended by \cite{zhang2016tpami} with corrected annotations. The Tsinghua-Daimler Cyclist (TDC) dataset \cite{Li2016} focuses on cyclists and other riders. In \cite{hwang2015multispectral} a multi-spectral dataset for pedestrian detection is introduced, combining RGB and infrared modalities. 

The Cityscapes dataset \cite{Cordts2016Cityscapes} was recorded in 50 cities during three seasons. Similar to earlier scene labeling challenges like Pascal VOC \cite{Everingham15} and Microsoft COCO \cite{lin2014microsoft}, it provides pixel-wise segmentations for a number of semantic object classes. The CityPersons dataset \cite{ben2017CityPerson} extends part of the Cityscapes dataset by bounding-box labels for the full extent of pedestrians. This enables occlusion analysis as the segmentation masks cover the visible areas only.

See Table \ref{tab:benchmarks_det} for an overview of the main person detection benchmarks in vehicle context. In terms of the annotation quantity and data diversity, CityPersons \cite{ben2017CityPerson} and Tsinghua-Daimler Cyclist \cite{Li2016} had, so far, the most to offer for the pedestrian and the riders class. Although Caltech \cite{dollar2012pedestrian} lists a large number of pedestrian annotations, only an unspecified subset of these annotations were done manually, the remainder was obtained by interpolation (we estimate the number of manual annotations to be an order of magnitude smaller). In total there are about 2300 unique persons in this dataset. Training and evaluation on Caltech is typically performed on a subset of the dataset, using every $30th$ frame. Cyclist and other riders annotations are missing in the Caltech dataset, and orientation annotations are missing in both Caltech and CityPersons datasets. KITTI, Caltech and TDC datasets have been collected in one city only. CityPersons was recorded in 27 different cities but, apart of Strasbourg and Zurich, it covers only Germany and recordings were not made throughout all seasons.
Very recently, the Berkeley Deep Drive dataset (BDD) \cite{bdd2018data} was made available, which in total provides $100000$ images recorded in a vehicle context. A white paper describing the dataset was announced.

Other person datasets outside the vehicle application are the attributes recognition datasets of Leibe \cite{sudowe2015person}, HATDB \cite{sharma2011learning}, Berkeley \cite{sermanet2013overfeat} and Hall \cite{hall2015fine}. They focus on the detection of body joints or the classification of attributes like male/female.
Notable for its sheer size is furthermore the very recent Open Images V4 dataset \cite{openimagesv4}, containing $15.4$M bounding boxes on $1.9$M images for $600$ different categories.

\begin{table*}[!htp]
\small
\vspace*{5pt}
\caption{Overview of recent deep learning detection methods.
Methods evaluated in this work are bold-faced.}
\label{tab:method_comparison}
\centering
\begin{tabular}{@{} r c c c c c c c @{}}
\toprule
& \multicolumn{3}{c}{two stage methods} & \phantom{} &  \multicolumn{3}{c}{one stage methods} \\
\cmidrule{2-4} \cmidrule{6-8}
& Fast R-CNN\cite{Girshick2015} & \textbf{Faster R-CNN}\cite{Ren2015} & \textbf{R-FCN}\cite{dai2016r} & & YOLO \cite{redmon2016you} & \textbf{YOLOv3} \cite{redmon2018} & \textbf{SSD}\cite{liu2016ssd} \\
\hline
region proposals              & external & RPN      & RPN  &&  gridbased & anchor boxes &  default boxes  \\
hard example mining      & implicit & implicit & explicit && none  & none & explicit \\
used feature maps & last  & last & last  && last & several & several \\
\bottomrule
\end{tabular}
\end{table*}

\subsection{Methods}

Deformable Part Models (DPM) using Histograms of Oriented Gradients (HOG) features \cite{Felzenszwalb2010,yan2013cvpr,ouyang2013single}, and Decision Forests using ICF features \cite{dollar2009integral,zhang2014,benenson2013,zhang2015filtered} were until a few years ago the established pedestrian detection methods \cite{benenson2014ten}.
Successes of deep learning for image classification (e.g. AlexNet \cite{krizhevsky2012imagenet}) also lead to its incorporation in object detection.
By training deep convolutional neural networks (CNN) like GoogleNet \cite{szegedy2015going}, VGG \cite{simonyan2014very} and ResNet \cite{he2016deep} on the ImageNet dataset for classification, models learn to extract powerful features from raw pixels, which can be used effectively for other tasks like object detection \cite{huh2016}.

A comparison of selected detection methods building up on feature maps of CNNs is shown in Table \ref{tab:method_comparison}.
They can be clustered into two stage methods \cite{Girshick2015} \cite{Ren2015} \cite{dai2016r}, that use a proposal stage and a downstream classification stage, and one stage methods that go without the proposal stage \cite{redmon2016you} \cite{redmon2018} \cite{liu2016ssd}.
The R-CNN methods \cite{Girshick2014}, \cite{Girshick2015}, \cite{Ren2015} are the basis for most current two stage methods.
R-CNN \cite{Girshick2014} and its extension Fast R-CNN \cite{Girshick2015} depend on  proposals for possible object locations from an external input. 
R-CNN uses a CNN to classify each proposal separately.
Fast R-CNN optimizes the runtime by executing the CNN on a complete image to share the calculated features.
For every (mapped) region proposal, features are pooled and used for separate classification and bounding box regression by fully connected layers.
The relation between proposal recall and the overall detection performance is shown in \cite{Hosang2016} for a lot of different proposal methods like selective search \cite{Uijlings2013},  MCG \cite{Arbelaez2014} and BING \cite{BingObj2014}.
Proposal methods based on depth data \cite{Chen2015, braun2016pose} increase the detection performance of Fast R-CNN as the proposal recall is larger.

Faster R-CNN \cite{Ren2015} does without external proposals by implementing a region proposal network (RPN).
Thus, the two stages are combined in a single network jointly trainable end-to-end.
Inside the RPN anchor-boxes of varying scales, positions, and aspect ratios are convolutionally classified  as fore- or background.
Foreground anchors are then used as proposals for feature pooling.
Regardless of the scale of an anchor-box only features are pooled from the last layer.
Hereby the spatial support of the features can be a lot larger or even smaller than the objects to be detected.
The problem of varying object sizes in pedestrian detection is tackled in the extensions \cite{LiLSXY15}, \cite{Cai2016ECCV}, \cite{Yang2016CVPR}, \cite{Zhu2016ACCV}.
In SDP \cite{Yang2016CVPR} features are pooled from different layers in dependence of the proposal size.
MS-CNN \cite{Cai2016ECCV} directly appends proposal networks on feature maps of different scales.

A great part of the computational complexity of Fast R-CNN and Faster R-CNN depends on the number of proposals.
The minibatches during training consist of a sampled subset, which is usually several orders of magnitude smaller than the total amount of proposals.
\cite{Girshick2015} and \cite{Felzenszwalb2010} argue that the selection of background samples slightly overlapping with positive samples can be seen as a heuristic hard negative mining.
R-FCN \cite{dai2016r} does not use fully connected layers and thus does not have to resort to limiting the number of proposals by sampling. Instead it uses convolutional layers to generate scoring maps.
Final detection is performed by pooling from these scoring maps without any further calculations dependent on trainable weights.
As all proposals are classified, online hard example mining \cite{ShrivastavaGG16} is applicable.

One stage detection methods like YOLO \cite{redmon2016you}, its extensions YOLOv2 \cite{redmon2017cvpr}, YOLOv3 \cite{redmon2018}, and others \cite{liu2016ssd}, \cite{Ren17CVPR} go without a distinct proposal stage.
In YOLO the final downsampled feature map is divided into grid cells.
For each grid cell fully connected layers are trained to detect objects that are centered within this cell using the complete image as spatial support.
This approach has weaknesses for small objects and object groups, that cluster within a single cell.
That is why YOLOv2 \cite{redmon2017cvpr} adopts the anchor boxes of Faster R-CNN.
Scales and aspect ratios of these boxes are set by calculating dimension clusters using k-means clustering.
Features are stacked from different layers to further support the detection of varying object sizes, still the boxes themselves are anchored in a single layer.
In YOLOv3 \cite{redmon2018} three different layers with three different strides are used to predict classes and precise positions for the anchor boxes.
Furthermore, they propose the Darknet-53 network architecture specialised for fast object detection, combining ideas of other CNNs \cite{szegedy2015going},\cite{simonyan2014very}, \cite{he2016deep}.

SSD \cite{liu2016ssd} detects objects based on default boxes.
These default boxes are similar to anchor boxes, but they are applied on different feature layers at different resolutions.
Hereby the receptive field sizes are approximately proportional to the sizes of the default boxes.
In the SSD512 variant, seven layers are used for prediction which means a finer discretization of the output space than with YOLOv3.
Unlike the YOLO methods not all negative boxes or gridcells are used in backpropagation.
Hard negative mining is applied to select the boxes with the highest confidence loss similar to R-FCN.
\cite{Ren17CVPR} introduces a recurrent neural network based on a VGG-16 architecture that improves the localisation accuracy of one stage methods.
This is achieved by applying a recurrent rolling convolution on several feature layers.

Generative adversarial networks (GAN) \cite{goodfellow2014generative} are also used for pedestrian detection.
In \cite{li2017cvpr} a Fast R-CNN architecture is extended by a generator branch that adds super resolved features after region proposal pooling to improve the detection performance for small objects.
The adversarial branch is trained to discriminate super resolved features of small objects from real features of large scale objects.
In \cite{huang2017cvpr} inspired by GANs a discriminator is trained to select realistic looking images rendered by a game engine.
An extension of Faster R-CNN coined RPN+ is then trained on this data to improve the detection performance for unusual pedestrians.

Deep learning has also been used for estimating orientations of common objects in traffic scenarios on datasets \cite{Geiger2012} that provide orientation ground-truth.
In \cite{Tulsiani2015} and \cite{Su2015}, orientation estimation is handled as a multi-class classification problem. 
\cite{Beyer2015} introduced the Biternion Net, which regresses continuous orientation angles. 
The Biternion representation is adapted in the Pose-RCNN \cite{braun2016pose} approach, such that orientation estimation is trained jointly with detection in a Fast R-CNN architecture.
In \cite{Chen2015} a L1 loss is used instead of the Biternion-based Von-Mises loss for estimating continuous orientation angles.
%
%
%
%
\subsection{Performance Analysis}
In \cite{dollar2012pedestrian}, 16 different detection methods are evaluated on the Caltech dataset. Small sizes and occlusion are identified as major challenges for pedestrian detectors. The reasonable set usually used for evaluation only contains pedestrians larger than 50 \textit{px} with no partial occlusion. In \cite{benenson2014ten} more than 40 detectors are evaluated on the Caltech dataset to analyze the main cause for improvement during the last 10 years. Deep models are examined as one of several possible causes. Still, they are outclassed by the design of better features as the main driver of performance improvement. In \cite{Hosang2015DnnForPedestrians} also deep models on the Caltech dataset are analyzed.
False positives which are touching ground-truth samples are considered as localization error. The remaining false positives are considered as confusion of background and foreground. Hereby, the authors find that confusion is the most frequent reason for false positives.
Discriminating false positives by localization and confusion errors is also done in \cite{zhang2016tpami}. The authors focus on the boosted decision forests-based methods RotatedFilters \cite{zhang2016cvpr} and Checkerboards \cite{zhang2015filtered}. In addition to categorizing false positives as localization or classification errors, they automatically analyze the effect of contrast, size and blurring on the detection score. Furthermore, they manually cluster false positives and false negatives at a fixed false positives per image by qualitative failure reasons. In contrast, \cite{rajaram2015exploration} applies an automatic failure analysis for ACF \cite{dollar2014pyramids} on Caltech and KITTI. They assign failure reasons to false negatives, such as truncation, occlusion, small objects heights, unusual aspect ratios, and localization in one study. As more than one of the sources could qualify as failure reason a certain prioritization provides the primary reason.

Methods \cite{Zhang2016,LiLSXY15,Cai2016ECCV} building upon the work of Fast/Faster R-CNN are the top-performing methods on the Caltech dataset \cite{zhang2016tpami}. \cite{Zhang2016} uses decision forests for classification instead of fully connected layers but the performance depends on the feature layers of the CNNs.
Regarding the KITTI benchmark, the top performing non-anonymous submissions all rely on deep CNNs \cite{Ren17CVPR,Cai2016ECCV,Xiang2016,Zhu2016ACCV,Yang2016CVPR}. Apart from \cite{Ren17CVPR} all of these are two stage methods building upon the work of Fast/Faster R-CNN.

\subsection{Main Contributions}

Our contributions are threefold:
\begin{itemize}
\item We introduce the EuroCity Persons dataset, which provides a large number of highly diverse, accurate and detailed annotations of persons (pedestrians, cyclists, and other riders) in urban traffic scenes across Europe. It also contains night-time scenes. Annotations extend beyond bounding boxes and include overall body orientations and a variety of object- and image-related tags. See Section \ref{sec:Benchmark}.  
\item We optimize four deep learning approaches (Faster R-CNN \cite{Ren2015}, R-FCN \cite{dai2016r}, SSD \cite{liu2016ssd} and YOLOv3 \cite{redmon2018}) to serve as baselines for the new person detection benchmark. We prove the generalization capabilities of detectors trained with the new dataset and thereby its usefulness. See Sections \ref{sub:baseline_experiments} and \ref{sub:Generalization}.

\item We provide insights regarding to the effect of several dataset characteristics on detector performance: the training set size, the dataset bias (day- vs. night-time, geographical region), the dataset detail (i.e. availability of object orientation information) and the annotation quality. We analyze error sources and discuss the road ahead. See Sections \ref{sub:DatasetAspects} and \ref{sec:Discussion}.
\end{itemize}

\section{Benchmark}
\label{sec:Benchmark}

\begin{figure*}
\begin{center}
    \includegraphics[width=0.33\linewidth]{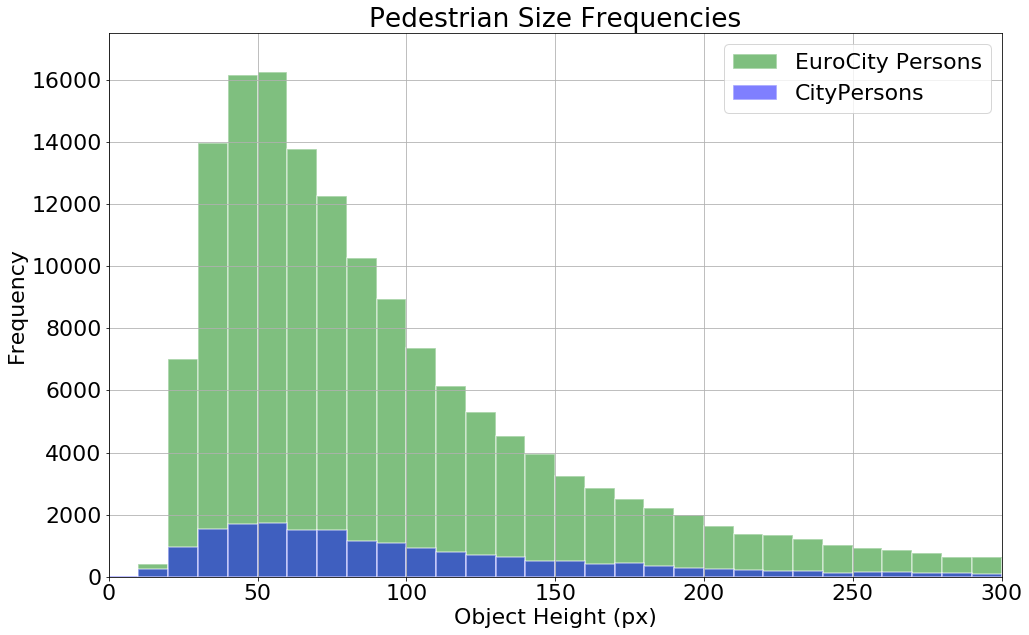}
    \includegraphics[width=0.33\linewidth]{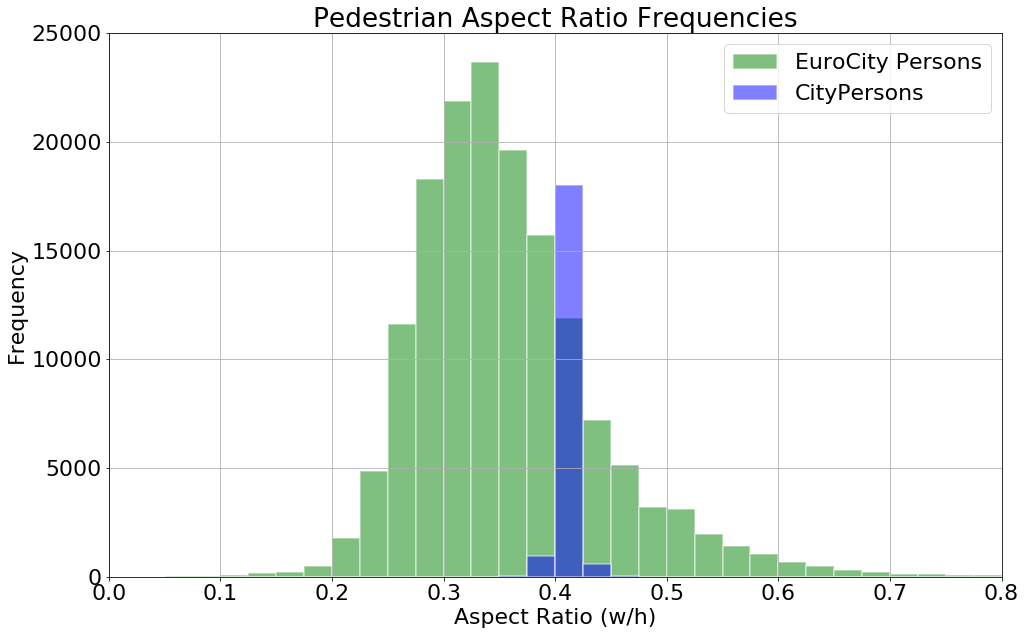}
    \includegraphics[width=0.33\linewidth]{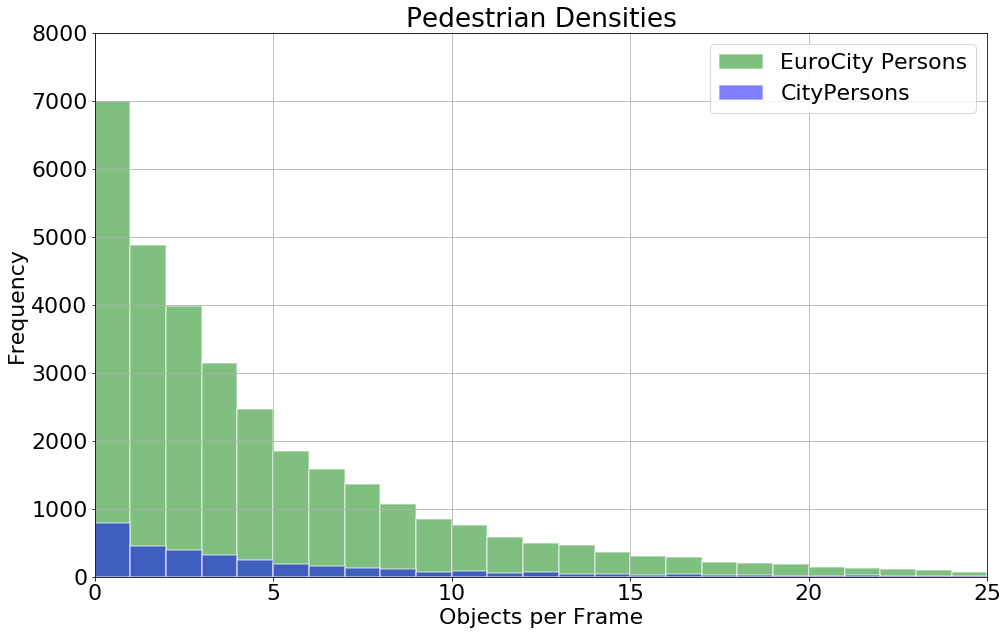}
\end{center}
    \caption{Statistics of EuroCity Persons and CityPersons for pedestrians of the training and validation datasets (height, aspect ratio and density).}
\label{fig:statistics}
\end{figure*}

\subsection{Dataset Collection}
We collected the images of the EuroCity Persons dataset from a moving vehicle in 31 cities of 12 European countries. Recordings were made with a state-of-the-art automotive-grade two megapixel camera (1920 x 1024) with rolling shutter at a frame rate of 20 Hz.
The camera, mounted behind the windshield, originally yielded 16 bit color images; this high dynamic-range was important for capturing scenes with strong illumination variation (e.g. night-time, low-standing sun shining directly into the camera). Images were debayered and rectified afterwards. For the purpose of EuroCity Persons benchmark, and for allowing comparisons with existing datasets, the original 16-bit color images were converted to 8-bit by means of a logarithmic compression curve with a parameter setting different for day and night. 

We collected 53 hours of image data in total, for an average of 1.7 hours per city. To limit selection bias \cite{torralba2011}, we extracted every 80-th frame for our detection benchmark without further filtering. This means that a substantial fraction of the person annotations in the dataset are unique, although especially at traffic lights and in slow moving traffic, same persons might appear in different annotations. Even so, due to sparse sampling at every four seconds, image resolutions and body poses will differ.

\subsection{Dataset Annotation}
\label{sub:dataset-annotation}

We annotated pedestrians and riders; the latter were further distinguished by their ride-vehicle type: bicycle, buggy, motorbike, scooter, tricycle, wheelchair.

\paragraph{Location} All objects were annotated with tight bounding boxes of the complete extent of the entity. If an object is partly occluded, its full extent was estimated (this is useful for later processing steps such as tracking) and the level of occlusion was annotated. We discriminated between no occlusion, low occlusion (10\%-40\%), moderate occlusion (40\%-80\%), and strong occlusion (larger than 80\%). Similar annotations were performed with respect to the level of object truncation at the image border (here, full object extent was not estimated). For riders, we labeled the riding person and its ride-vehicle with two separate bounding boxes, and annotated the ride-vehicle type. Riderless-vehicles of the same type in close proximity were captured by one class-specific group box (e.g. several bicycles on a rack).

In \cite{zhang2016tpami} and \cite{ben2017CityPerson} one vertical line is drawn and automatically converted into a rectangular box of a fixed aspect ratio. Because of the diverse pedestrian aspect ratios (see Figure \ref{fig:statistics} middle) and to be comparable with the KITTI dataset, we remained with the classical bounding-box convention of labeling the outermost object parts. 
For every sampled frame, all visible persons were annotated; otherwise, missed annotations could lead to the flawed generation of background samples during training and bootstrapping. Also persons in non-upright poses (e.g. sitting, lying) were annotated or persons behind glass. These cases were tagged separately.

A person is annotated with a rectangular (class-specific) ignore region if a person is smaller than 20 \textit{px}, if there are doubts that an object really belongs to the appropriate class, and if instances of a group can not be discriminated properly. In the latter case, several instances may be grouped inside a single ignore region. 

\paragraph{Orientation} The overall object orientation is an important cue for the prediction of future motion of persons in traffic scenes. We provide this information for all persons larger than 40 \textit{px} (including those riding). 

\paragraph{Additional Tags} Person depictions (e.g. large poster) and reflections (e.g. in store windows) were annotated as a separate object class. Additional events were tagged at the image level, such a lens flare, motion blur, and rain drops or a wiper in front of the camera.

All annotations were manually performed; no automated support was used, as it might introduce an undesirable bias towards certain algorithms during benchmarking. 
We placed reasonably high demands on accuracy. The amount of missed and hallucinated objects were each to lie within 1\% of the annotated number. Annotators were asked to be accurate within two pixels for bounding box sides (apart from ignore regions) and within 20 degrees for orientation.
Annotations were double checked by a quality validation team that was disjoint from the annotation team. If needed, several feedback iterations were run between the teams to achieve a consolidated outcome. Experiments regarding annotation quality are listed in Section \ref{sub:DatasetAspects}.

\subsection{Data Subsets}
\label{subsec:data_subsets}

We define various data subsets on the overall EuroCity Persons dataset. First, we distinguish a day-time and a night-time data subset, each with its own separate training, validation and test set. 
Three overlapping data subsets are furthermore defined, considering the ground-truth annotations, similar to \cite{Geiger2012, Li2016, ben2017CityPerson}:
\begin{itemize}
\item \textbf{Reasonable:~}Persons with a bounding box height greater than 40 \textit{px} which are occluded/truncated less than 40\%
\item \textbf{Small:~} Persons with a height between 30 \textit{px} and 60 \textit{px} which are occluded/truncated less than 40\%
\item \textbf{Occluded:~} Persons with a bounding box height greater than 40 \textit{px} which are occluded between 40\% and 80\%
\end{itemize}
These data subsets can be used to selectively evaluate properties of person detection methods for various sizes or degrees of occlusion.

\begin{figure}[t]
\begin{center}
   \includegraphics[width=0.98\linewidth]{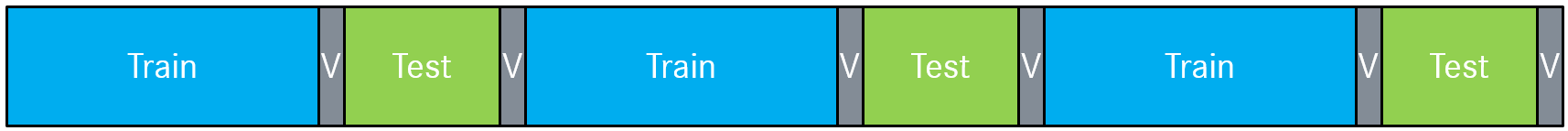}
\end{center}
   \caption{The applied test, val, and train split visualized for one city. Assuming a recording length of one hour for this city, the whole session is divided into three equidistant $20$ minute subsets. Each subset is then split into train, validation, and test by a $60\%, 10\%, 30\%$ distribution.}
\label{fig:datasetsplit}
\end{figure}

Each city recording lasted on average 1.7 hours. In order to increase the chances that certain time-dependent environmental conditions (e.g. a rain shower, particular type of road infrastructure or buildings) were well represented across training, validation and test set, for each city the recordings are separated into chunks with a duration of at least $20$ minutes. The recorded images of each chunk were split into training, evaluation, and test by $60\%$, $10\%$, and $30\%$ respectively, as illustrated in Figure \ref{fig:datasetsplit}. During halts due to traffic lights or jams people could appear in several consecutive frames. To facilitate that the test, validation and training sets are disjunct in terms of people we only splitted sequences at points in time where the recording vehicle had a speed larger than 7 $km/h$. By placing furthermore the validation set intermittently with the training and test, it was all but avoided that the latter two would contain the same physical person.  

\subsection{Dataset Characteristics}
\label{subsec:dataset_characteristics}

See Table \ref{tab:benchmarks_det} and Figure \ref{fig:statistics} for some statistics on the new EuroCity Persons dataset. Seasonality, weather, time of day and, to some degree, geographical location, all influence clothing and thus person appearance. These factors also influence the person density observed, which, as shown in Figure \ref{fig:statistics} (right) varies a lot, not only per frame but also per city. For example, the lowest average number of pedestrians per city (1.8) occurred in Leipzig likely due to the rainy weather during recording. Very crowded scenarios have been collected in Lyon with on average 9.5 pedestrians per image. These imply challenging occlusions and overlapping objects that complicate non-maximum suppression (these difficult scenarios are missing in KITTI and Caltech, where on average there is about one pedestrian per frame). Geographical location also influences the background (i.e. vehicles, road furniture, buildings). The time-of-the-day has furthermore a significant impact on scene appearance. Recordings at night-time suffer from low contrast, color loss and motion blur.

By driving through a large part of Europe, during all four seasons, in most weather conditions (apart from heavy rain or snowfall), and during day and night, we recorded very diverse backgrounds and person appearances, see Table \ref{tab:qualitative_true_positives}.

\subsection{Evaluation Metrics}
\label{subsec:evaluation_protocol}

To evaluate detection performance, we plot the miss-rate ($mr$) against the number of false positives per image ($fppi$) in log-log plots:
\begin{align}
mr(c)   &= \frac{fn(c)}{tp(c) + fn(c)},\\
fppi(c) &= \frac{fp(c)}{\#img},
\end{align}
where $tp(c)$ is the number of true positives, $fp(c)$ is the number of false positives, and $fn(c)$ is the number of false negatives, all for a given confidence value $c$ such that only detections are taken into account with a confidence value greater or equal than $c$.
As commonly applied in object detection evaluation \cite{Geiger2012, dollar2012pedestrian, Everingham15, ben2017CityPerson} the confidence threshold $c$ is used as a control variable. 
By decreasing $c$, more detections are taken into account for evaluation resulting in more possible true or false positives, and possible less false negatives.
We define the log average miss-rate ($LAMR$) as
\begin{align}
LAMR = \exp{\left(\frac{1}{9} \sum_{f} \log{\left( mr( \argmax_{fppi(c) \leq f} fppi(c) ) \right)} \right)},
\end{align}
where the $9$ $fppi$ reference points $f$ are equally spaced in the log space, such that $f \in \{10^{-2}, 10^{-1.75}, \dots, 10^0\}$.
For each $fppi$ reference point the corresponding $mr$ value is used.
In the absence of a miss-rate value for a given $f$ the highest existent $fppi$ value is used as new reference point, which is enforced by $mr( \argmax_{fppi(c) \leq f} fppi(c) )$.
This definition enables $LAMR$ to be applied as a single detection performance indicator at image level. At each image the set of all detections is compared to the ground-truth annotations by utilizing a greedy matching algorithm. An object is considered as detected (true positive) if the Intersection over Union (\textit{IoU}) of the detection and ground-truth bounding box exceeds a pre-defined threshold. Due to the high non-rigidness of pedestrians we follow the common choice of an \textit{IoU} threshold of $0.5$. Since no multiple matches are allowed for one ground-truth annotation, in the case of multiple matches the detection with the largest score is selected, whereas all other matching detections are considered false positives. After the matching is performed, all non matched ground-truth annotations and detections, count as false negatives and false positives, respectively.
In addition, to allow a comparison with results from other work \cite{Geiger2012, Li2016} we also utilize the Average Precision (AP), which is defined as:
\begin{align}
AP = \frac{1}{11} \sum_{r \in \{0, 0.1, \dots, 1 \}} \max_{re(c) \geq r} pr(c),
\end{align}
with the recall $re(c) = tp(c) / (tp(c) + fn(c))$, and precision $pr(c) = tp(c) / (tp(c) + fp(c))$, both for a given confidence threshold $c$. 

For the evaluation of joint object detection and pose estimation we use the average orientation similarity (AOS) \cite{Geiger2012}:
\begin{align}
AOS = \frac{1}{11} \sum_{r \in \{0, 0.1, \dots, 1 \}} \max_{\tilde{r}:\tilde{r}\geq r} s(\tilde{r}),
\end{align}
where $s$ is the orientation similarity given by:
\begin{align}
s(r) = \frac{1}{|\mathcal{D}(r)|} \sum_{i \in \mathcal{D}(r)} \frac{1 + \cos \Delta^{(i)}_\theta}{2}\delta_i.
\end{align}
$\mathcal{D}(r)$ denotes the set of all object detections at recall $r$ and $\Delta^{(i)}_\theta$ is the difference between the estimated and the ground-truth angle. $\delta_i$ is set to 1, if detection $i$ has been assigned to a ground truth bounding box ($IoU > 0.5$) else it is set to zero, to penalize multiple detections which explain a single object. Thus, the upper bound of the $AOS$ is given by the $AP$ score. 

As in \cite{Geiger2012}, \cite{ben2017CityPerson}, neighboring classes and ignore regions are used during evaluation. Neighboring classes involve entities that are semantically similar, for example bicycle and moped riders. Some applications might require their precise distinction (\textbf{enforce}) whereas others might not (\textbf{ignore}). In the latter case, during matching correct/false detections are not credited/penalized. If not stated otherwise, neighboring classes are ignored in the evaluation.
In addition to ignored neighboring classes all persons annotations with the tags \textit{behind glass} or \textit{sitting-lying} are treated as ignore regions. Further, as mentioned in Section \ref{sub:dataset-annotation}, ignore regions are used for cases where no precise bounding box annotation is possible (either because the objects are too small or because there are too many objects in close proximity which renders the instance based labeling infeasible).
Since there is no precise information about the number or the location  of objects in the ignore region, all unmatched detections which share an intersection of more than $0.5$ with these regions are not considered as false positives.

\subsection{Benchmarking}
\label{subsec:benchmarking}

The EuroCity Persons dataset, including its annotations for the training and validation sets, is made freely available to academic and non-profit organizations for non-commercial, scientific use. The test set annotations are withheld. An evaluation server is made available for researchers to test their detections, following the metrics discussed in previous Subsection. Results are tallied online, either by name or anonymous. The frequency of submissions is limited.

\section{Experiments}
\label{sec:Experiments}

All the baseline- and generalization experiments (Sections \ref{sub:baseline_experiments} and Section \ref{sub:Generalization}) involved the day-time EuroCity Persons dataset and the pedestrian class, for comparison purposes with earlier works. This also holds in part for the data aspects experiments (Section \ref{sub:DatasetAspects}), unless stated otherwise.

\subsection{Baselines}
\label{sub:baseline_experiments}

As the top ranking methods on KITTI and Caltech use deep convolutional neural networks, we select our baselines among these methods. Many recent pedestrian detection methods \cite{li2017cvpr}, \cite{huang2017cvpr}, \cite{mao2017cvpr}, \cite{Cai2016ECCV}, \cite{Xiang2016}, \cite{Zhu2016ACCV}, \cite{Yang2016CVPR} are extensions of Fast/Faster R-CNN and profit from the basic concepts of these methods. Therefore, \textbf{Faster R-CNN} is evaluated as prominent representative of the two stage methods. As shown in \cite{ben2017CityPerson}, it can reach top performance for pedestrian detection if it is properly optimized.
The one stage methods often trade faster inference against a lower detection accuracy. YOLO \cite{redmon2016you} is one of the first methods within this group. We evaluate its latest extension \textbf{YOLOv3} \cite{redmon2018}, as in comparison with its predecessors, its design is promising regarding the detection of smaller objects.
Within both groups we also select methods with explicit hard example mining, namely \textbf{R-FCN} \cite{dai2016r} and \textbf{SSD} \cite{liu2016ssd}.

For Faster R-CNN, R-FCN and SSD we use VGG-16 \cite{simonyan2014very} as base network as it is very common in pedestrian detection \cite{ben2017CityPerson},\cite{Ren17CVPR}, \cite{li2017cvpr}, \cite{huang2017cvpr}, \cite{mao2017cvpr}.
YOLOv3 is trained with the Darknet framework \cite{darknet} and Darknet-53 \cite{redmon2018} as base architecture whereas the other three methods are trained with Caffe \cite{jia2014caffe} and VGG-16 \cite{simonyan2014very}.
The base networks are pre-trained on ImageNet.

\begin{figure}[t]
\begin{center}
  \includegraphics[width=0.9\linewidth]{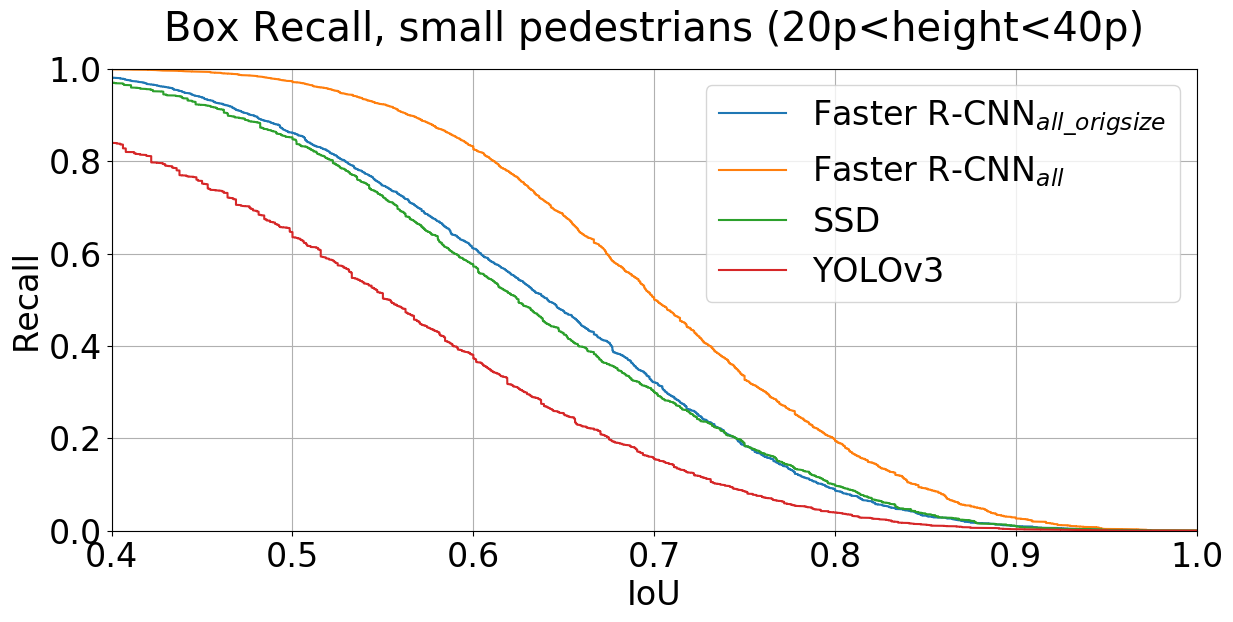}
  \includegraphics[width=0.9\linewidth]{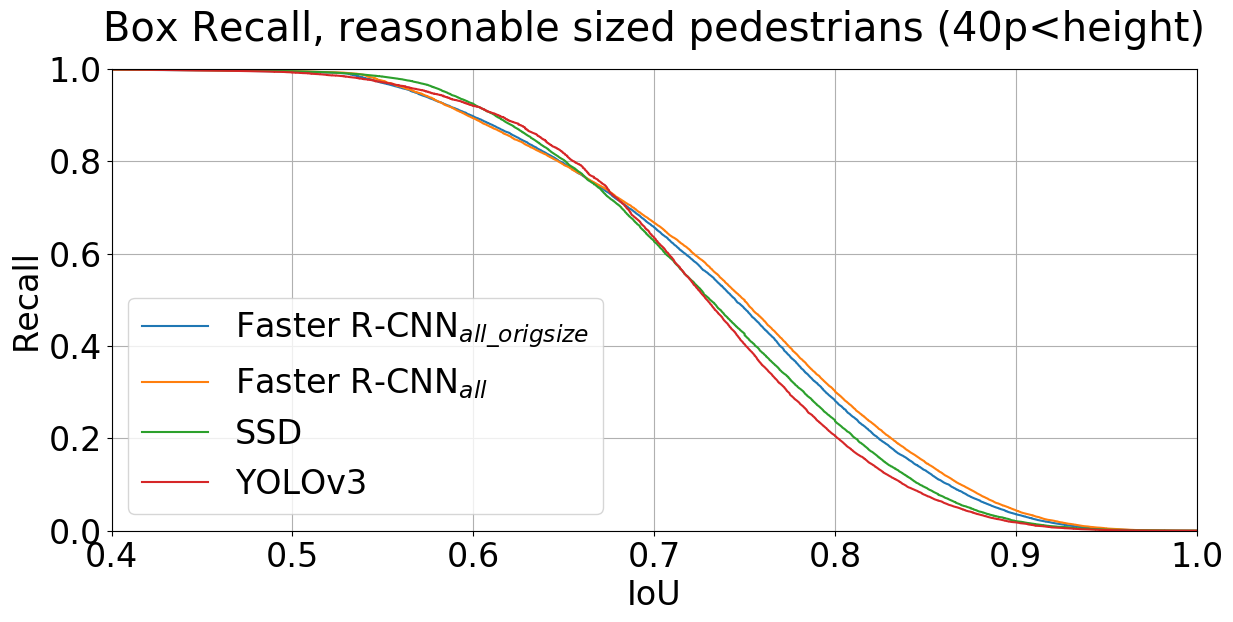}
\end{center}
    \caption{Recall vs. \textit{IoU} for small pedestrians (top) and pedestrians of the reasonable scenario (down) for the optimized anchor-boxes of Faster R-CNN and YOLOv3 and the SSD default boxes.}
\label{fig:proposalrecall}
\end{figure}

\paragraph{Box Recall Improvements}
All four methods evaluated have in common that an object without any matching anchor box or default box can neither be used for training nor detected during testing.
Therefore we first optimize the box recall. 
For Faster R-CNN and R-FCN we apply some improvements from \cite{ben2017CityPerson} adapting the scales and aspect ratios of the anchor-boxes, reducing the feature stride by removing the last max pooling layer and upscaling the input image during training and testing.
Since the annotation protocol employed for our dataset does not enforce a fixed aspect ratio for all objects, we utilized two different aspect ratios which were empirically chosen according to Figure \ref{fig:statistics} (i.e. median value of $0.34 \pm 0.5$). 
Furthermore, eight different anchor-box scales were used, to capture objects across all sizes.
Lastly, to improve performance for small sized objects the input images are upscaled by a constant factor (see Table\ref{tab:fasterrcnn_improvements} for the R-CNN variants, R-FCN is also upscaled the same way), resulting in a greater number of anchor-boxes. In our experiments, the available GPU memory of 12 GB limits the maximal upscaling factor. SSD and YOLOv3 can thus not be trained with upscaled images. For SSD we add additional convolutional layers as in the SSD512 model \cite{liu2016ssd} to achieve a greater spatial support for larger objects. The two selected aspect ratios for the anchor boxes are also used for the default boxes.

For YOLOv3 we cluster the pedestrian sizes within the training dataset into nine anchor box sizes as described in \cite{redmon2017cvpr}.
Every of the three prediction layers is responsible for the detection based on three of the nine anchor box sizes.

The resulting box recalls on the training dataset are shown in Figure \ref{fig:proposalrecall}. The recall for pedestrian boxes of the reasonable scenario larger than 40 \textit{px} is about 100\% for all methods for an \textit{IoU} of 0.5. Boxes of Faster R-CNN$_{all}$ are evaluated on $1.3 \times$ upscaled images in contrast to Faster R-CNN$_{all\_origsize}$ resulting in a higher recall in particular for smaller pedestrians.

\paragraph{Further Improvements}
We implement an ignore region handling for Faster R-CNN and R-FCN, as in \cite{ben2017CityPerson}. 
As explained before, ignore regions might contain objects of a given class without precise localization.
The ignore region handling prevents the sampling of background boxes in those areas, that could potentially overlap with real objects.

Training with very small or strongly occluded samples could lead to models detecting a lot more false positives. 
For the reasonable test scenario, only middle sized and moderately occluded pedestrians have to be detected.
Hence, we filter training samples according to different test scenarios to train several Faster R-CNN models.
The filter and upscaling settings are summarized in Table \ref{tab:fasterrcnn_improvements}.
Filtered objects are handled as ignore regions by Faster R-CNN and R-FCN in order to ensure that they are not sampled as background during training.
For all experiments with R-FCN, SSD and YOLOv3 we filter samples that are more than 80\% occluded or smaller than 20 \textit{px} in height.

\paragraph{Training Strategy and Model Selection}
We use SGD as backpropagation algorithm for all our trainings and stepwise reduce the learning rate.
The training progress is observed on the validation dataset.
For Faster R-CNN and R-FCN the learning rate is initialized with $10^{-3}$, reduced to $10^{-4}$ after 100k iterations and further reduced to $10^{-5}$ after 250k iterations. 
Training is finished after 270k iteration.
For YOLOv3 and SSD we reduce the initial learning rate to $10^{-4}$ to avoid exploding losses. 
Both methods also require a longer training time based on the slower training progress.
For YOLOv3 (SSD) we reduce the learning rate to $10^{-5}$ after 400k (1000k) iterations and to $10^{-6}$ after 450k (1300k) iterations.
Training is finished after 500k (1400k) iterations.
At the end of training only the best snapshot on the validation dataset is selected and evaluated on the test dataset, to avoid optimizing on the latter.

\begin{table}[!htp]
\small
\vspace*{5pt}
\caption{Training settings of the Faster R-CNN method, differing in the heights and degree of occlusion of the samples used for training and in the upscaling factor used by bilinear interpolation (between brackets).}
\label{tab:fasterrcnn_improvements}
\centering
\begin{tabular}{@{} l c c c  @{}}
\toprule
                               & height & occlusion & upscaling \\
\hline
Faster R-CNN$_{small}$            & $[20, \infty]$ & $[0, 40]$ & yes (1.3)\\
Faster R-CNN$_{reasonable}$       & $[40, \infty]$ & $[0, 40]$ & yes (1.3)\\
Faster R-CNN$_{occluded}$         & $[40, \infty]$ & $[0, 80]$ & yes (1.3)\\
Faster R-CNN$_{all}$              & $[20, \infty]$ & $[0, 80]$ & yes (1.3)\\
Faster R-CNN$_{all\_origsize}$    & $[20, \infty]$ & $[0, 80]$ & no\\
Faster R-CNN$_{baseline}$         & $[20, \infty]$ & $[0, 40]$ & no\\
\bottomrule
\end{tabular}
\end{table}

\paragraph{Results}
See Table \ref{tab:results} for the quantitative results obtained with the methods considered. Variants of the two stage method Faster R-CNN perform overall best on the three test scenarios.
Faster R-CNN$_{small}$ performs best on the corresponding small test scenario, and interestingly also slightly better on the reasonable setup.
Faster R-CNN$_{all}$, that is trained with pedestrians of all sizes and of occlusions up to 80\%, performs best overall. 
It also performs slightly better than Faster R-CNN$_{occluded}$ on the occluded test scenario.
The Faster R-CNN variants ($all\_origsize, baseline$) that are trained and tested with the original image resolution perform slightly worse for the reasonable and occluded test scenario than the other Faster R-CNN variants. 
Still, they run 66\% faster during training and testing. 
As could be expected by the lower box recall shown in Figure \ref{fig:proposalrecall}, there is a considerable performance difference for small sized pedestrians.
Interestingly, both one stage detectors YOLOv3 and SSD perform better than R-FCN at least on the reasonable and occluded test scenarios.
One of the main differences between Faster R-CNN and R-FCN is the use of the bootstrapping method OHEM. 
OHEM proves useful when comparing results for the two R-FCN variants with enabled and disabled OHEM for the occluded scenario.

See Table \ref{tab:qualitative_true_positives} for some illustrations of typical results with Faster R-CNN$_{all}$ (we include night-time and rider results, not part of this section).

\begin{table}[!htp]
\small
\vspace*{5pt}
\caption{Log average miss-rate ($LAMR$) on the test set of the EuroCity Persons benchmark for different settings of the optimized methods.}
\label{tab:results}
\centering
\begin{tabular}{@{} l c c c  @{}}
& \multicolumn{3}{c}{Test Scenario} \\
\cmidrule{2-4}
                               & reasonable & small & occluded \\
\hline
Faster R-CNN$_{small}$            & \textbf{7.3} & \textbf{16.6} & 52.0 \\
Faster R-CNN$_{reasonable}$       & 7.4 & 23.4 & 50.8 \\
Faster R-CNN$_{occluded}$         & 7.9 & 24.1 & 34.2 \\
Faster R-CNN$_{all}$              & 8.1 & 17.1 & \textbf{33.9} \\
Faster R-CNN$_{all\_origsize}$    & 9.4 & 23.4 & 35.4 \\
Faster R-CNN$_{baseline}$         & 9.4 & 22.8 & 55.1 \\
YOLOv3                            & 8.5 & 17.8 & 37.0 \\  
SSD                               & 10.5 & 20.5 & 42.0 \\
R-FCN OHEM                        & 12.1 & 19.6 & 44.0 \\
R-FCN NoOHEM                      & 12.2 & 19.5 & 45.6 \\

\bottomrule
\end{tabular}
\end{table}

\begin{figure*}
\begin{center}
    \includegraphics[width=0.33\linewidth]{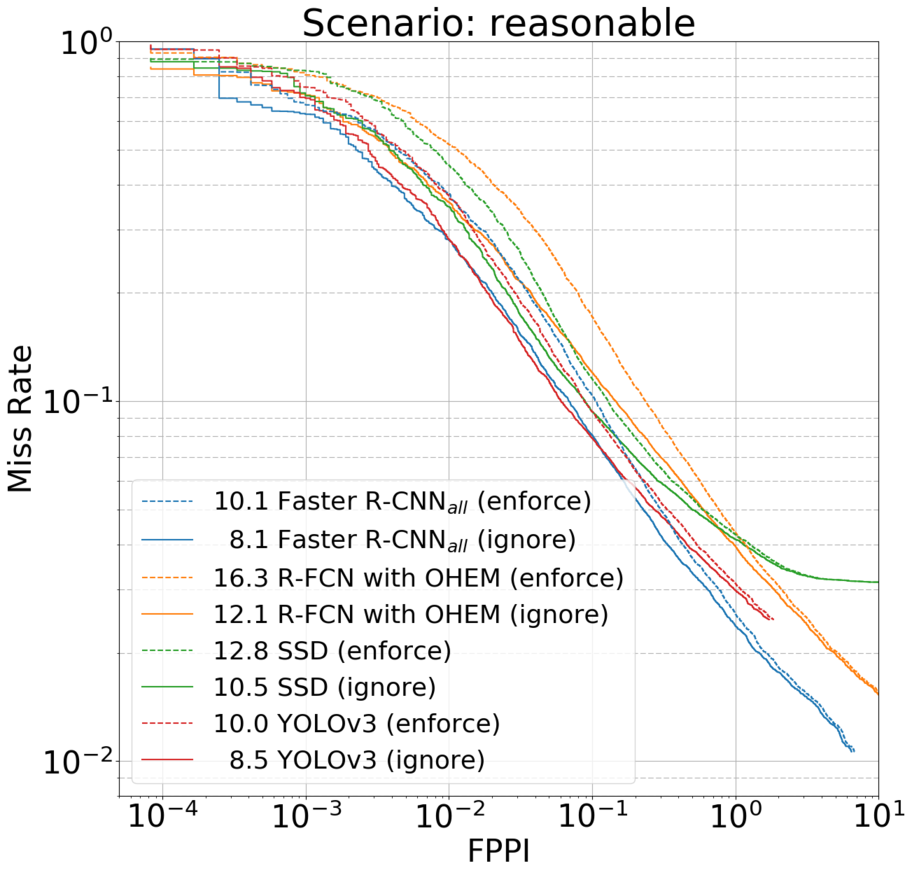}
    \includegraphics[width=0.33\linewidth]{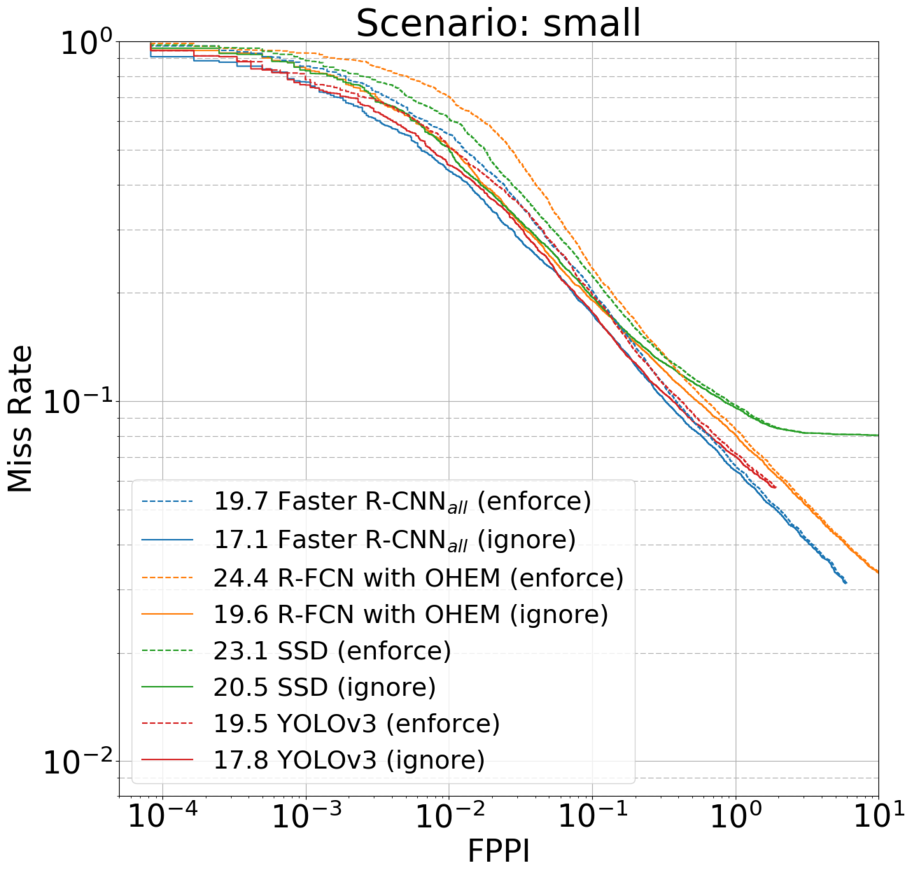}
    \includegraphics[width=0.33\linewidth]{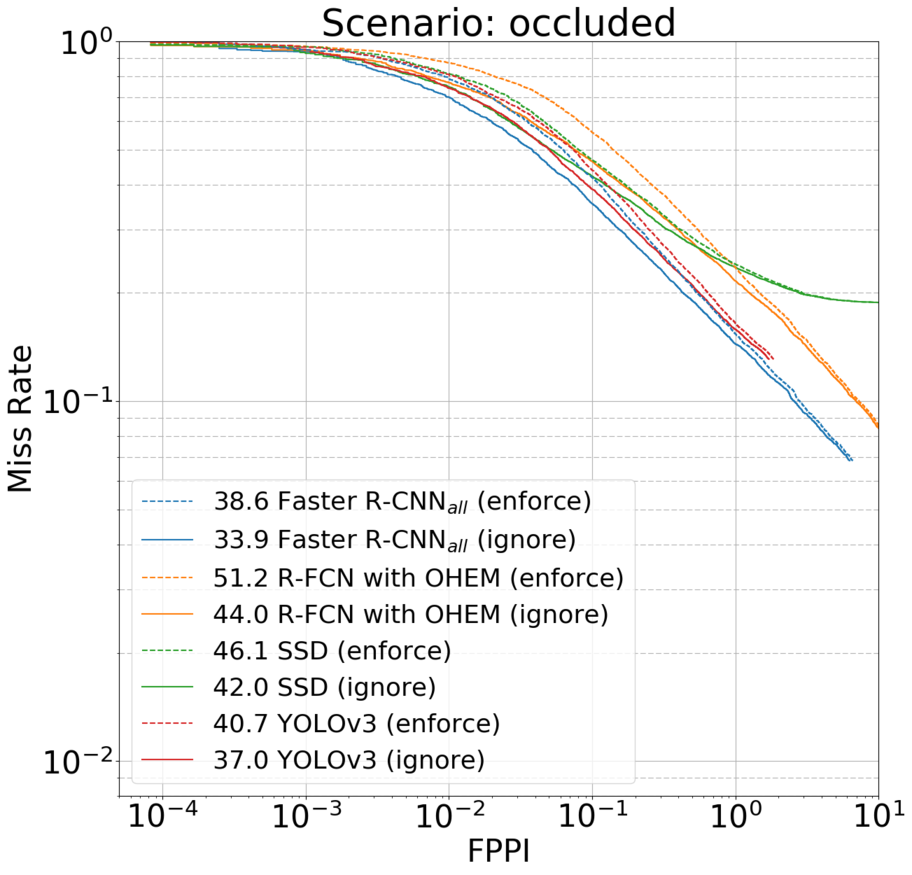}
\end{center}
    \caption{Miss-rate curves on the EuroCity Persons test set for our selected methods for the reasonable (left), small (middle) and occluded (right) test setting. The required \textit{IoU} for a detection to be matched with a ground-truth sample is 0.5. For every method, the curves are shown for enforcing or ignoring precise class label with respect to neighboring classes.}
\label{fig:missrate_curves}
\end{figure*}

\paragraph{Failure Analysis}
We now analyze the detection errors of our best-performer on Faster R-CNN$_{all}$ qualitatively and quantitatively. Table \ref{tab:qualitative_mistakes} illustrates false positives and false negatives of this method at a false positive per image rate of 0.3 for the reasonable test scenario, clustered by main error source. As can be seen, clothes, depictions and reflections are main sources for confusion with real pedestrians and thus for false positives (our evaluation policy is strict and we count these wrong due to application considerations; note, however, that depictions and reflections are annotated in our dataset, thus a more lenient policy to ignore false positives of these type is readily implemented).

Certain pedestrian poses and aspect ratios can lead to multiple detections for the same pedestrian as shown in the \textit{Multidetections} category.
Non-maximum suppression (NMS) is used by Faster R-CNN and other deep learning methods to suppress multiple detections. We use an \textit{IoU} threshold of 0.5 which is not sufficient to suppress detections that have very diverse aspects. On the other hand, a higher \textit{IoU} threshold would result in more false negatives.
These already occur for an \textit{IoU} threshold of 0.5 as shown in the \textit{NMS repressing} category.
In these scenarios, pedestrians are occluded less than 40\% and thus have to be detected in the reasonable test scenario.
Because of the high \textit{IoU} between pedestrians not all of them can be detected because of the greedy NMS. 
Thus, NMS is an important part of many deep learning methods that is usually not trained but has a great influence on detection performance.

Small and occluded pedestrians are a further common source for false negatives as already shown by the \textit{small} and \textit{occluded} test scenarios.
These two groups have also been analyzed before \cite{rajaram2015exploration}.
In street scenarios usually only the lower part of a pedestrian is occluded due to parked cars or other obstacles.
In our qualitative analysis we have false negatives where the head is occluded.
These are particularly challenging for pedestrian detection methods, as these cases are quite rare in the training dataset.
Further challenges are rare poses or pedestrians leaning on bicycles as shown in the \textit{Others} group.

\begin{figure}
\begin{center}
    \includegraphics[width=0.8\linewidth]{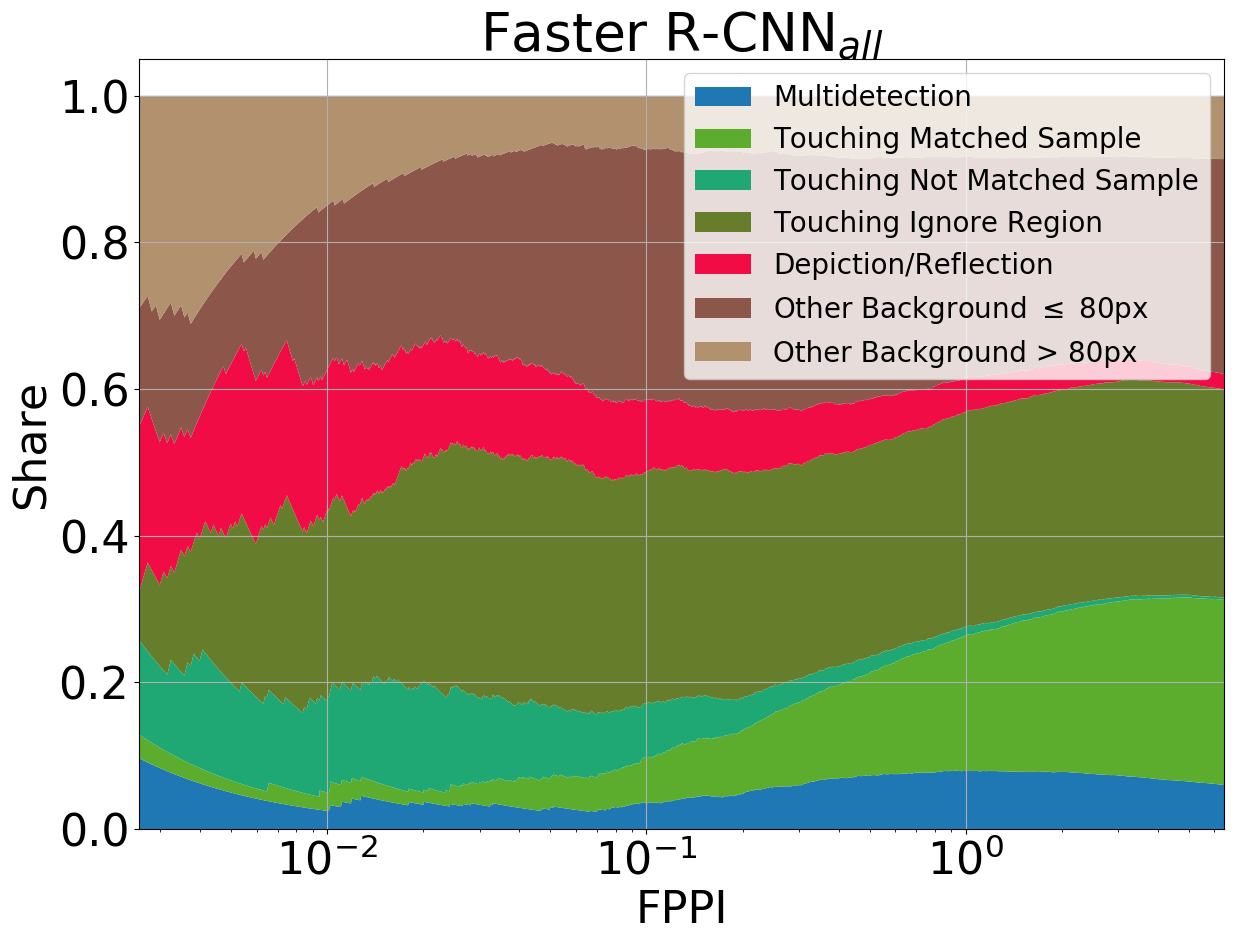}
\end{center}
    \caption{The contribution of various sources to the number of false positives of Faster R-CNN$_{all}$, depending on $fppi$}
\label{fig:quantitative_failure_analysis}
\end{figure}

For the quantitative analysis of false positives we build upon the ideas of oracle tests as in \cite{zhang2016tpami}. There, false positives touching ground-truth samples are regarded as localization error. Non-touching false positives are regarded as confusion of fore- and background.
We analyze false positives types for a finely discretized range of false positive per image ($fppi$), see Figure \ref{fig:quantitative_failure_analysis}.
In this study, we further subdivide the localization errors in four groups: multiple detections ($IoU > 0.5$ with ground-truth samples, as we penalize multiple assignments), and detections touching matched ground truth samples, non-matched ground truth samples, and ignore regions, respectively. In this context an ignore region may either be an ignore region annotation or an object that has not to be detected in the reasonable test scenario. We also subdivide the fore- and background confusions into three groups: detections that can be matched with depictions and reflections, and other background, further subdivided whether smaller than 80 \textit{px} or not.

Figure \ref{fig:quantitative_failure_analysis} shows that localization errors account for about 60\% of all errors at a high $fppi$ of 6, decreasing to about 40\% for a low $fppi$ rate of $4 \times 10^{-3}$. The share of false positives touching ground-truth samples remains approximately the same for the entire $fppi$ range. Of these touched ground-truth samples, an increasing proportion is non-matched, for decreasing $fppi$. The share of false positives touching ignore regions is similar for a large $fppi$ range but decreases somewhat for $fppi$ below $10^{-2}$. Possible objects inside these ignore regions seem to lead to erroneous detections in their surroundings. In terms of classification errors, depictions and reflections are among the hardest error sources to take care off: at decreasing $fppi$ the share of this error type increases. Also the share of larger other-background objects increases with decreasing $fppi$.

\newcommand\colsize{1.65cm}
\newcommand\multicolsize{4.3cm}

\begin{table*}
	\setlength{\tabcolsep}{0.2em}
     \renewcommand{\arraystretch}{1.2}
  \caption{Qualitative detection results of Faster R-CNN$_{all}$ at $fppi$ of 0.3 (green: pedestrians, blue: riders). Samples are recorded during dry weather (first row), rainy weather and wintertime (second row), and during dusk and night (last two rows).} 

    \begin{tabular}{c c c}
   \hline 
			      \multicolumn{3}{l}{\textbf{True Positives}} \\
     \hline
   \includegraphics[width=0.33\linewidth]{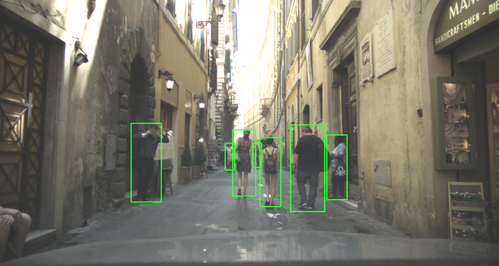} 
   & \includegraphics[width=0.33\linewidth]{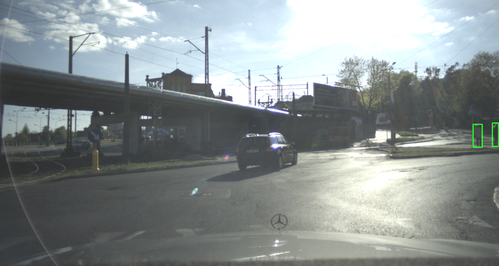} 
   & \includegraphics[width=0.33\linewidth]{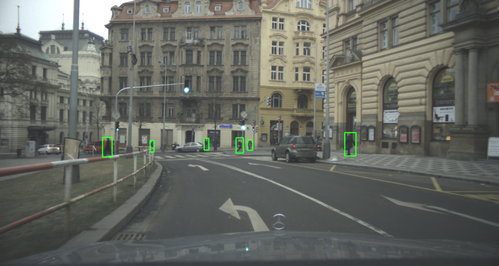} \\

   \includegraphics[width=0.33\linewidth]{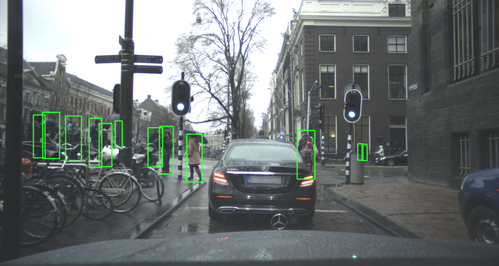} 
   & \includegraphics[width=0.33\linewidth]{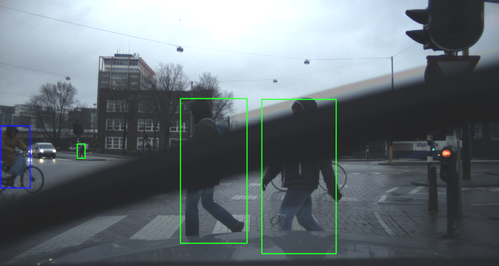} 
   & \includegraphics[width=0.33\linewidth]{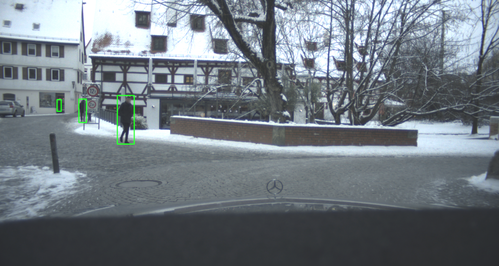} \\
		
   \includegraphics[width=0.33\linewidth]{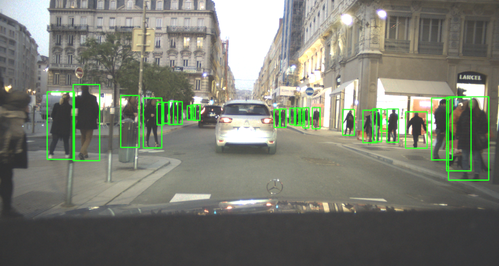} 
   & \includegraphics[width=0.33\linewidth]{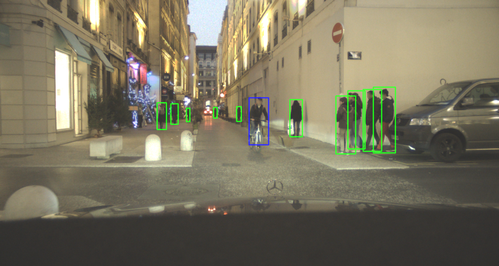} 
   & \includegraphics[width=0.33\linewidth]{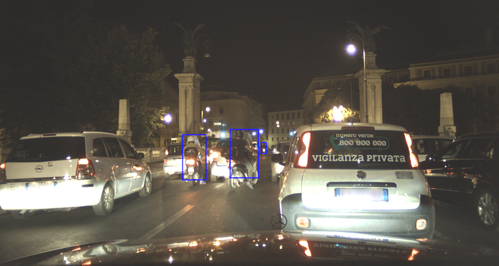} \\	

   \includegraphics[width=0.33\linewidth]{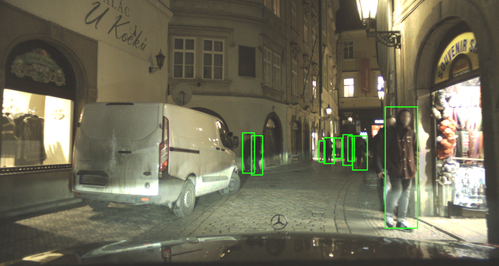} 
   & \includegraphics[width=0.33\linewidth]{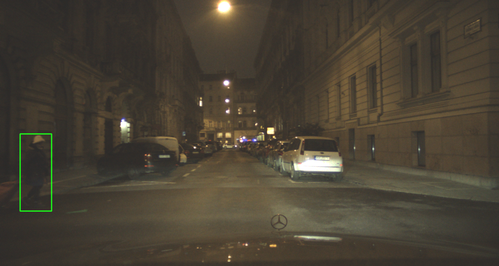} 
   & \includegraphics[width=0.33\linewidth]{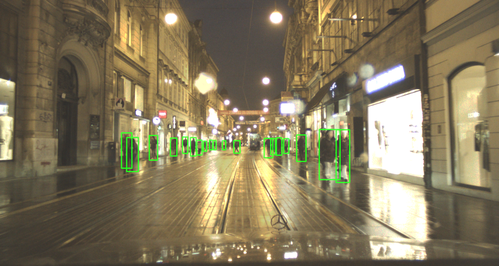} \\
    \end{tabular}
\label{tab:qualitative_true_positives}
\end{table*}

\begin{table*}
	\setlength{\tabcolsep}{0.2em}
     \renewcommand{\arraystretch}{1.2}
  \caption{Qualitative detection results for Faster R-CNN$_{all}$ at 0.3 $fppi$ (green: true positives, red: false positives, purple: false negatives, white: ground truth).}

  \begin{tabular}{p{0.2cm}p{\colsize}p{\colsize}p{\colsize}p{\colsize}p{\colsize}p{0.2cm}p{\colsize}p{\colsize}p{\colsize}p{\colsize}p{\colsize}}
      \hline 
      \multicolumn{3}{l}{\textbf{False Positives (Image Detail)}} \\
      \hline 
      \rotatebox[origin=c]{90}{Clothes}
      & \parbox[c]{1em}{\includegraphics[width=\colsize]{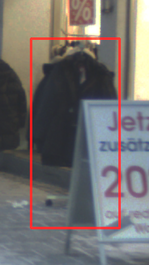}} 
    & \parbox[c]{1em}{\includegraphics[width=\colsize]{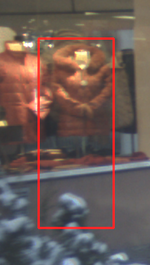}} 
      & \parbox[c]{1em}{\includegraphics[width=\colsize]{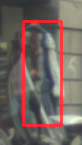}} 
      & \parbox[c]{1em}{\includegraphics[width=\colsize]{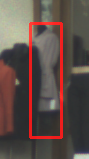}} 
      & \parbox[c]{1em}{\includegraphics[width=\colsize]{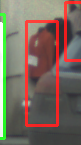}} 
      & \rotatebox[origin=c]{90}{Background}
      & \parbox[c]{1em}{\includegraphics[width=\colsize]{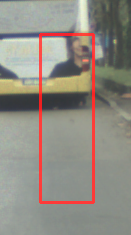}} 
      & \parbox[c]{1em}{\includegraphics[width=\colsize]{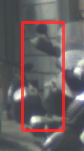}} 
      & \parbox[c]{1em}{\includegraphics[width=\colsize]{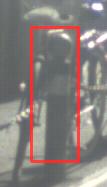}} 
      & \parbox[c]{1em}{\includegraphics[width=\colsize]{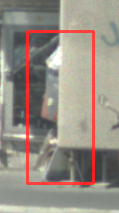}}
      & \parbox[c]{1em}{\includegraphics[width=\colsize]{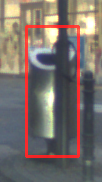}}  \\ 
      \rotatebox[origin=c]{90}{Labelerror}
      & \parbox[c]{1em}{\includegraphics[width=\colsize]{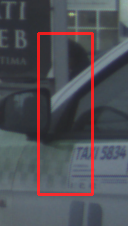}} 
    & \parbox[c]{1em}{\includegraphics[width=\colsize]{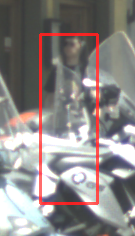}} 
      & \parbox[c]{1em}{\includegraphics[width=\colsize]{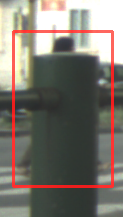}} 
      & \parbox[c]{1em}{\includegraphics[width=\colsize]{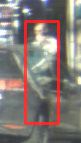}}
      & \parbox[c]{1em}{\includegraphics[width=\colsize]{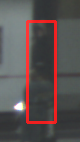}}
      & \rotatebox[origin=c]{90}{Depiction}
      & \parbox[c]{1em}{\includegraphics[width=\colsize]{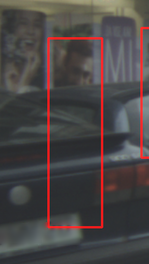}} 
      & \parbox[c]{1em}{\includegraphics[width=\colsize]{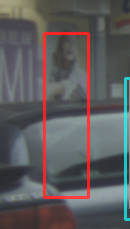}} 
      & \parbox[c]{1em}{\includegraphics[width=\colsize]{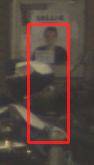}} 
      & \parbox[c]{1em}{\includegraphics[width=\colsize]{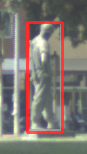}} 
      & \parbox[c]{1em}{\includegraphics[width=\colsize]{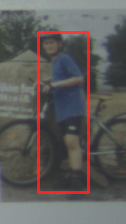}}  \\ 
      \rotatebox[origin=c]{90}{Multidetections} 
      & \parbox[c]{1em}{\includegraphics[width=\colsize]{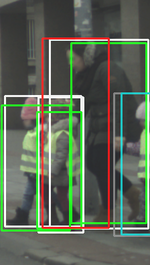}} 
    & \parbox[c]{1em}{\includegraphics[width=\colsize]{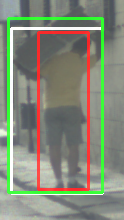}} 
      & \parbox[c]{1em}{\includegraphics[width=\colsize]{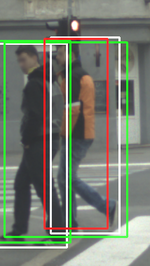}} 
      & \parbox[c]{1em}{\includegraphics[width=\colsize]{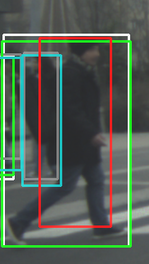}}
      & \parbox[c]{1em}{\includegraphics[width=\colsize]{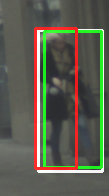}} 
      & \rotatebox[origin=c]{90}{Reflection}
      & \parbox[c]{1em}{\includegraphics[width=\colsize]{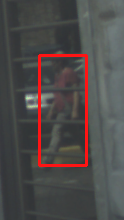}} 
      & \parbox[c]{1em}{\includegraphics[width=\colsize]{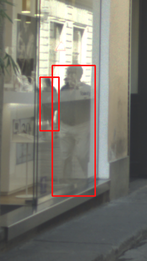}} 
      & \parbox[c]{1em}{\includegraphics[width=\colsize]{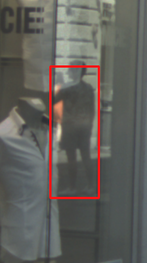}} 
      & \parbox[c]{1em}{\includegraphics[width=\colsize]{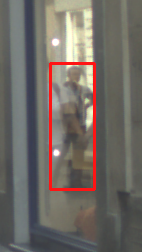}}  
      & \parbox[c]{1em}{\includegraphics[width=\colsize]{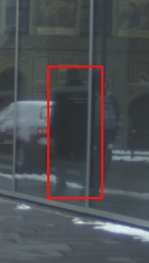}} \\
      
      \hline 
       \multicolumn{3}{l}{\textbf{False Negatives (Image Detail)}} \\
      \hline
           \rotatebox[origin=c]{90}{Small size}
      & \parbox[c]{1em}{\includegraphics[width=\colsize]{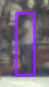}} 
    & \parbox[c]{1em}{\includegraphics[width=\colsize]{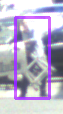}} 
      & \parbox[c]{1em}{\includegraphics[width=\colsize]{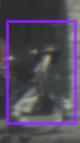}} 
      & \parbox[c]{1em}{\includegraphics[width=\colsize]{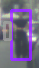}}       
      & \parbox[c]{1em}{\includegraphics[width=\colsize]{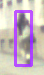}}
      & \rotatebox[origin=c]{90}{Occlusion}
      & \parbox[c]{1em}{\includegraphics[width=\colsize]{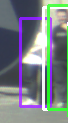}}
      & \parbox[c]{1em}{\includegraphics[width=\colsize]{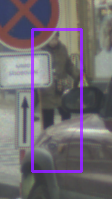}}       
      & \parbox[c]{1em}{\includegraphics[width=\colsize]{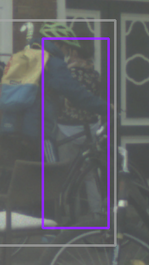}} 
      & \parbox[c]{1em}{\includegraphics[width=\colsize]{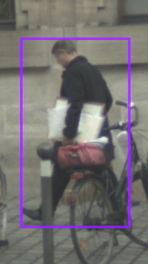}} 
      & \parbox[c]{1em}{\includegraphics[width=\colsize]{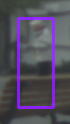}}  \\
      \rotatebox[origin=c]{90}{NMS repressing}
      & \parbox[c]{1em}{\includegraphics[width=\colsize]{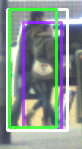}}  
    & \parbox[c]{1em}{\includegraphics[width=\colsize]{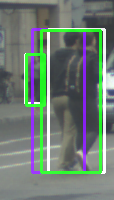}} 
      & \parbox[c]{1em}{\includegraphics[width=\colsize]{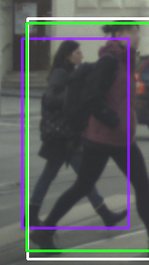}}     
      & \parbox[c]{1em}{\includegraphics[width=\colsize]{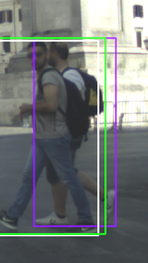}} 
      & \parbox[c]{1em}{\includegraphics[width=\colsize]{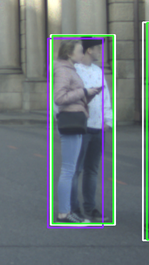}}
      & \rotatebox[origin=c]{90}{Others}
      & \parbox[c]{1em}{\includegraphics[width=\colsize]{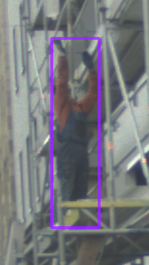}}
      & \parbox[c]{1em}{\includegraphics[width=\colsize]{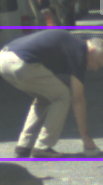}}
      & \parbox[c]{1em}{\includegraphics[width=\colsize]{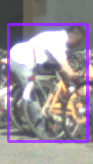}} 
      & \parbox[c]{1em}{\includegraphics[width=\colsize]{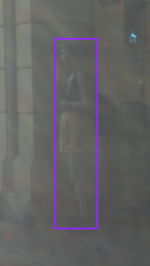}}
      & \parbox[c]{1em}{\includegraphics[width=\colsize]{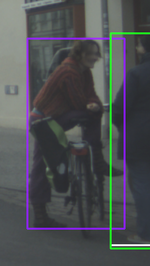}}  \\
    \end{tabular}
\label{tab:qualitative_mistakes}
\end{table*}

\begin{figure*}
\begin{center}
    \includegraphics[width=0.33\linewidth]{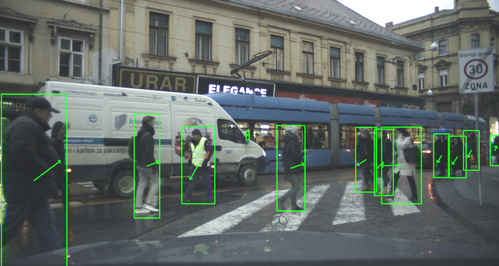}
	    \includegraphics[width=0.33\linewidth]{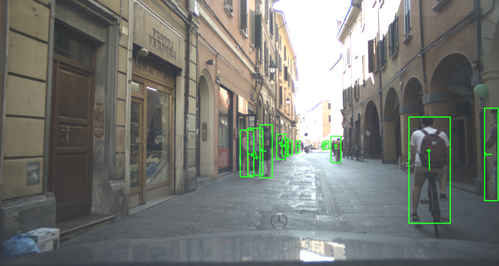}
			\includegraphics[width=0.33\linewidth]{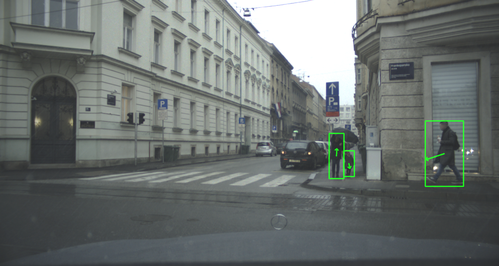}	 
\end{center}
    \caption{Qualitative results for orientation estimation. Left and middle image show correct estimations. Right image contains a rare failure case (left person has orientation offset of about 180 degrees)}
\label{fig:qualitative_orientation_estimation}
\end{figure*}

\paragraph{Computational Efficiency}
Processing rates for the R-FCN, Faster R-CNN, SSD and YOLOv3 on non-upscaled test images were 1.2 $fps$, 1.7 $fps$, 2.4 $fps$ and 3.8 $fps$, respectively, on a Intel(R) Core(TM) i7-5960X CPU \@ 3.00 GHz processor and a NVidia GeForce GTX TITAN X with 12.2 GB memory. There are several possibilities to optimize the runtime, such as replacing the VGG base architecture by a GoogLeNet model \cite{szegedy2015going} and upgrading to the latest GPU processor; this was outside the scope of this study.

For our remaining experiments we focus on Faster R-CNN as best performing method. Results for other methods are shown when they lead to additional insights.

\subsection{Generalization Capabilities}
\label{sub:Generalization}

A dataset with a reduced bias should better capture the true world, and result in superior generalization capabilities of the detectors which are trained on this dataset. KITTI and EuroCity Persons for example differ in camera types used for recording. Even for a human the images of these datasets look differently regarding colors and style.
The CityPersons (CP) and EuroCity Persons (ECP) datasets have been recorded with similar cameras. Still they differ regarding annotation bias, as the aspect ratios of all bounding boxes provided by CP are the same in contrast to ECP (cf. Section \ref{sub:dataset-annotation}).
Because of these inherent dataset biases models trained on one dataset have to be fine-tuned to be applicable on other datasets.
Thus, to demonstrate the increased diversity of our dataset we train and test our optimized Faster R-CNN baseline method on the two datasets, KITTI and CP with and without first pre-training on ECP.
For KITTI we split the official training dataset into two equally sized, disjunct subsamples used for training and validation as in \cite{Chen2015}.

The results on the validation sets are shown for the KITTI dataset in Table \ref{tab:results_pretrain_kitti} and for the CP dataset in Table \ref{tab:results_pretrain_cp}.
The overall detection performance on both datasets is superior for the cases in which ECP is used for pre-training.
These findings not only hold on the validation datasets.
Faster R-CNN 
achieved an average precision of 65.9 for pedestrians on the official KITTI test set for the moderate scenario. Our model pre-trained on ECP achieved an average precision of 72.6 on the KITTI test set.
Despite the dataset biases, the models were able to learn general features for the task of pedestrians detection when pre-trained on ECP which proves useful for other datasets as well after finetuning.
On the other hand, when using CP to pretrain for KITTI and vice versa, the detection performance is worse than when using ECP for pre-training.
For the moderate setting of the KITTI benchmark the Faster R-CNN pretrained on CP achieves an average precision of $77.5$.
The model pre-trained on ECP outperforms the CP based model with an average precision of $80.8$.

The same findings hold for the CP benchmark, on which pre-training on KITTI data has no influence on the detection performance that remains at a $LAMR$ of $17.2$. Again, the best results were achieved by the Faster R-CNN model pre-trained on ECP with a $LAMR$ of $14.9$. Note that the $LAMR$ listed in \cite{ben2017CityPerson} for training and testing on CP was 12.8 rather than 17.2 listed here. The difference arises from a difference in reasonable settings used. If we use the exact same settings as in \cite{ben2017CityPerson}, we arrive at an even better $LAMR$ of 12.2, which is improved by ECP pre-training to 10.9. 

The superior performance on both KITTI and CP datasets, when using ECP for pre-training indicates a high dataset diversity; models trained on this dataset will have increased generalization capabilities. However, due to the dataset biases solely training on a dataset from the other domain without fine-tuning results in worse detection performance.
Using transfer learning to pretrain a network on generic data and fine-tune on the target domain is widely applied and used to increase overall performance \cite{yosinski2014transferable, razavian2014cnn}.

\begin{table}[!htp]
\small
\vspace*{5pt}
\caption{Average Precision on the KITTI validation set for different training settings of Faster R-CNN. $A \rightarrow B$ denotes pre-training on $A$ and finetuning on $B$.}
\label{tab:results_pretrain_kitti}
\centering
\begin{tabular}{@{} l c c c  @{}}
& \multicolumn{3}{c}{KITTI Validation Set} \\
\cmidrule{2-4}
      Training Data                 & easy & moderate & hard \\
\hline
KITTI                             & 80.8 & 72.3 & 62.6 \\
ECP                                & 73.9 & 68.7 & 61.4 \\
ECP$\rightarrow$KITTI              & \textbf{85.6} & \textbf{80.8} & \textbf{73.9} \\
CP                                & 69.8 & 65.2 & 58.6 \\
CP$\rightarrow$KITTI              & 83.6 & 77.5 & 68.5 \\
\bottomrule
\end{tabular}
\end{table}

\begin{table}[!htp]
\small
\vspace*{5pt}
\caption{Log average miss-rate ($LAMR$) on the CityPersons (CP) validation set for different training settings of Faster R-CNN. $A \rightarrow B$ denotes pre-training on $A$ and finetuning on $B$.}
\label{tab:results_pretrain_cp}
\centering
\begin{tabular}{@{} l c c c  @{}}
& \multicolumn{3}{c}{CityPersons Validation Set} \\
\cmidrule{2-4}
     Training Data          & reasonable & small & occluded \\
\hline
CP                                & 17.2 & 38.9 & 52.0 \\
ECP                                & 22.5 & 41.2 & 53.7 \\
ECP$\rightarrow$CP                 & \textbf{14.9} & \textbf{30.0} & \textbf{44.8} \\
KITTI                             & 57.7 & 81.4 & 87.2 \\
KITTI$\rightarrow$CP              & 17.2 & 37.7 & 49.5 \\
\bottomrule
\end{tabular}
\end{table}

\subsection{Dataset Aspects}
\label{sub:DatasetAspects}
As stated in \cite{sun2017} pushing performance boundaries by using better datasets is not yet at its end.
Thus, they propose to focus on the datasets instead of applying more and more bells and whistles to recent methods resulting in the danger to overfit on existing datasets.
We argue that the generalization capabilities needed for transfer learning shown in the last section can be addressed to four dataset aspects, namely, diversity, quantity, accuracy, and detail.
By an extensive evaluation and artificially disturbance experiments we empirically show the importance of these aspects in the case of pedestrian detection in an automotive setup.
Faster R-CNN$_{baseline}$ is used as training setting without upscaling images because of computational considerations. 

\begin{figure}[t]
\label{fig:onecol}
\begin{center}
   \includegraphics[width=0.75\linewidth]{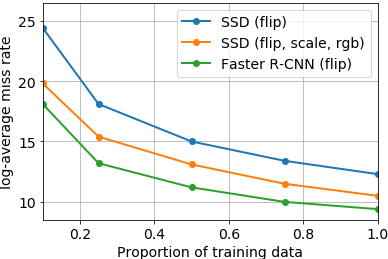}
\end{center}
   \caption{Detection performance ($LAMR$) of Faster R-CNN and SSD as a function of training set size}
\label{fig:exp_quantity}
\end{figure}

\paragraph{Quantity}
\cite{sun2017} shows a logarithmic relation between the amount of training data and the performance of deep learning methods. We validate this relation on our benchmark.
Therefore we train our baseline methods on different sized subsets which are randomly sampled from all cities.
The detection results for our baseline methods with the use of different augmentation modes in dependence of the dataset proportion are shown in Figure \ref{fig:exp_quantity}.
As image augmentations the images may be flipped or scaled in size.
The \textit{rgb} augmentation randomly shifts the colors of an image independently for the three color channels.
We observe that logarithmic relation between training set size and detection performance also holds on our benchmark for Faster R-CNN and SSD.

\paragraph{Diversity}
%
We wish to investigate whether overall geographical region introduces a dataset bias which influences person detection performance. For this, we constructed two datasets that are similar in terms of other influencing factors (i.e. season, weather, time of day, person density):
\begin{itemize}
\item \textit{Central West Europe (WE):} Basel, Dresden, K\"oln, N\"urnberg, Stuttgart, Ulm, W\"urzburg
\item \textit{Central East Europe (EE):} Bratislava, Budapest, Ljubljana, Prague, Zagreb
\end{itemize}
We split these datasets into subsets for training, validation and testing as described in section \ref{fig:datasetsplit}, such that the number of pedestrians in each training dataset is $15000$.
\cite{Demsar} shows that resampling of a dataset can be applied to evaluate the significance of benchmark results.
We permutate the train-val-test blocks and vary the block length (between $10$ and $30$ minutes) resulting in 20 different dataset combinations for training, validation and testing.
For every dataset combination one model is trained per region and evaluated on the corresponding test datasets of the two regions.
The mean performances over all different dataset combinations and the standard deviations for these are shown in Table \ref{tab:diversity_geolocation}.
In the case of a non existent dataset bias the difference between the output of both models comes from a distribution with zero median.
This is used as the null-hypothesis for the Wilcoxon signed-rank test \cite{Demsar}.
For the same test set the 20 results for the model trained on the same location and the model trained on the other location are paired.
We calculate the respective $p$-value, which is the probability of observing the test results given the null-hypothesis is true.
For the WE and EE test sets, these values are $0.0098$ and $0.0020$, respectively.
Hence, with a confidence interval of $99 \%$, the null-hypothesis (the non-existence of a regional bias) can be rejected for both regions.

Another diversity factor is the time of day.
Table \ref{tab:nighttime_daytime} shows detection results for the day-time, night-time and combined datasets. As the night-time dataset is only 20\% of day data (Table \ref{tab:benchmarks_det}) we reduce for this experiment the number of training samples used for the day-time and combined models accordingly. Table \ref{tab:nighttime_daytime} shows that training on day-time and testing on night-time gives significantly worse results than training and testing on the same time-of-day. Overall results are worse than those of other experiments due to the comparatively small training sets used.

\begin{table}[!htp]
\small
\vspace*{5pt}
\caption{Effect of geographical bias on detection performance ($LAMR$) for the reasonable scenario: central West Europe (WE) vs. central East Europe (EE). Datasets compiled to provide otherwise similar conditions. Results involve averages over different dataset splits.}

\label{tab:diversity_geolocation}
\centering
\begin{tabular}{@{} l c c c c @{}}
\toprule
& \multicolumn{4}{c}{Test Set} \\
\cmidrule{2-5}
               Training Set & WE (mean) & WE (std) & EE (mean) & EE (std)\\
\hline
WE                & 12.8 & 1.3    &  11.2        & 0.7 \\
EE                & 14.5 & 2.4	  & \textbf{9.2} & 0.5\\
WE\&EE			  & \textbf{12.4} & 1.4 & 9.8  & 0.8\\
\bottomrule
\end{tabular}
\end{table}

\begin{table}[!htp]
\small
\vspace*{5pt}
\caption{Effect of day- vs. night-time condition on detection performance ($LAMR$) for the reasonable scenario. Datasets compiled to provide otherwise similar conditions.}
\label{tab:nighttime_daytime}
\centering
\begin{tabular}{@{} l c c  @{}}
\toprule
& \multicolumn{2}{c}{Test Set} \\
\cmidrule{2-3}
               Training Set & Night & Day \\
\hline
Night                         & \textbf{18.6} & 21.3 \\
Day                           & 33.4 & \textbf{14.5} \\
Day and Night                       & 23.1 & 14.6 \\
\bottomrule
\end{tabular}
\end{table}

\paragraph{Detail}
The importance of additional annotations for ignore regions, for riders, and for orientations is now examined. Table \ref{tab:detail_ignore} shows results for a model trained without ignore region handling compared to our baseline method. In accordance with earlier findings \cite{ben2017CityPerson}, we observe that detection performance deteriorates when not using ignore regions during training. For the reasonable and small test scenario the $LAMR$ drops by about two points.

\begin{figure}[t]
\begin{center}
   \includegraphics[width=0.80\linewidth,height=0.60\linewidth]{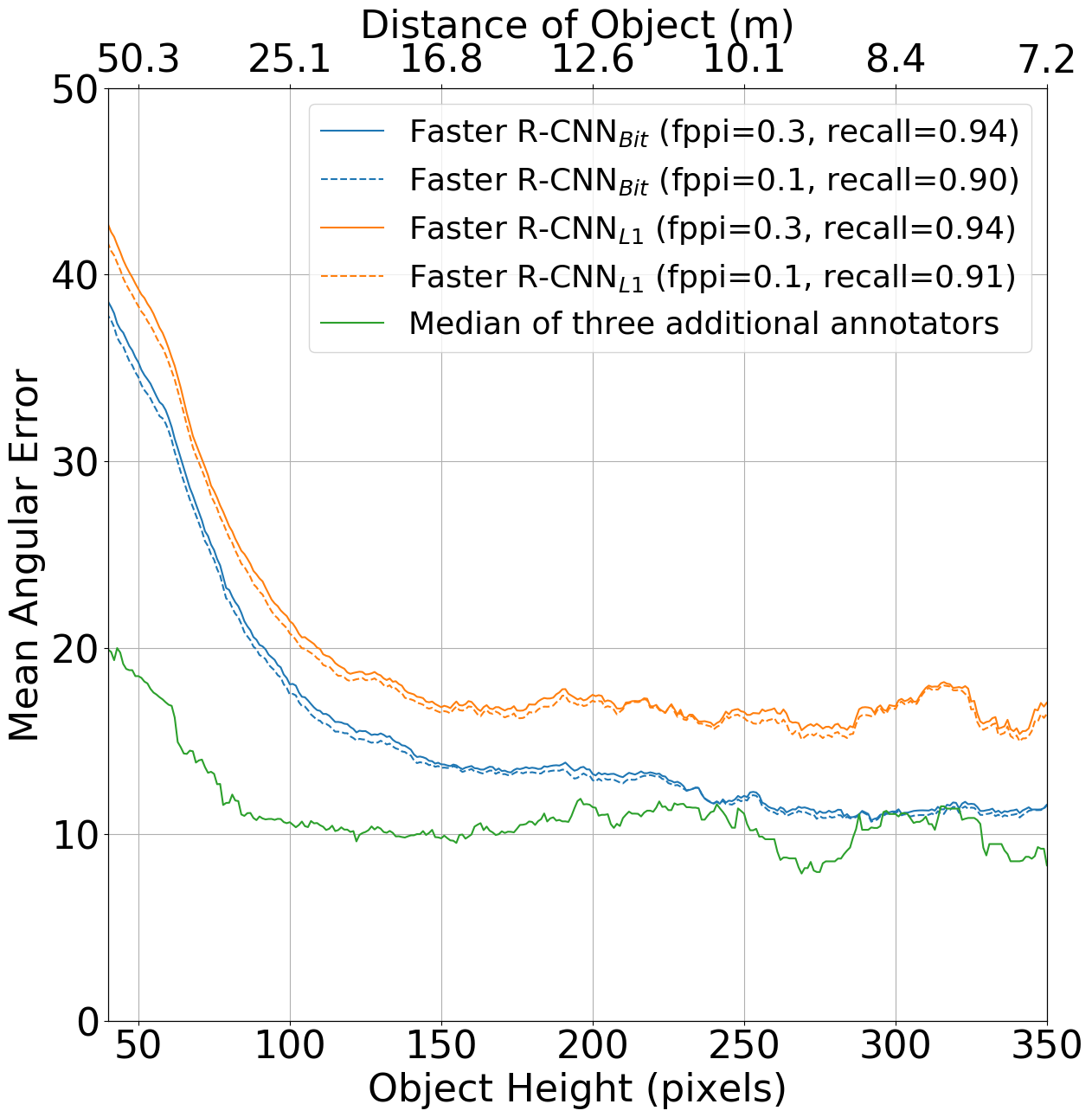}
\end{center}
\caption{Person orientation estimation quality vs. object size (distance).}
\label{fig:OrientationEstimation}
\end{figure}

We extended the baseline detection method by an orientation estimation layer as in \cite{braun2016pose} (Two variants for the orientation loss is considered: L1 and Biternion loss). Hence, the network performs multi-tasking: classification, bounding box and orientation regression. As body orientation correlates with the aspect ratio we assume that the bounding box regression task and hereby the detection performance could also benefit from learning all three tasks jointly in one network. In contrast to \cite{Girshick2015} which shows that training multiple tasks together can improve the overall result, Table \ref{tab:detail_ignore} shows that the detection results decrease slightly for the multitask network with the Biternion loss. 
Figure \ref{fig:OrientationEstimation} shows the orientation estimation error as a function of object size (distance). The Biternion loss is superior to the L1 loss as it does not suffer from the periodicity of an orientation angle. Using the aggregated AOS metric from Section \ref{subsec:evaluation_protocol} for the reasonable test scenario we get a score of 86.0 for the L1 loss and 86.7 for the Biternion loss.  

The evaluation protocol described in Section \ref{subsec:evaluation_protocol} ignores detected neighboring classes. For pedestrians this means that riders are not considered as false positives. If these neighboring classes are instead counted as false positives, detection performance decreases as expected: the $LAMR$ for our baseline method increases from 9.4 to 11.0, as shown in Table \ref{tab:detail_classes}. By adding riders as an additional class, one observes that the pedestrian detection performance improves for the protocol which requires pedestrians to be classified as such (10.3 vs. 11.0).
There is only a slight difference in performance when the network is trained to regress a bounding box for the rider alone or for the rider including the ride type.
The absolute detection performance for pedestrians and riders is quite similar although there are 10 times more pedestrians than riders in our training dataset.
\begin{table}[!htp]
\small
\vspace*{5pt}
\caption{Log average miss-rate ($LAMR$) of the detail study.}
\label{tab:detail_ignore}
\centering
\begin{tabular}{@{} l c c  @{}}
\toprule
& \multicolumn{2}{c}{Test Scenario} \\
\cmidrule{2-3}
               Training Scenario & reasonable & small \\
\hline
Baseline                           & \textbf{9.4} & \textbf{22.8} \\
NoIgnoreHandling                  & 10.9 & 24.5 \\
Orientation L1                    & \textbf{9.4} & 22.9 \\
Orientation Bit                   & 10.1 & 24.1 \\

\bottomrule
\end{tabular}
\end{table}

\begin{table}[!htp]
\small
\vspace*{5pt}
\caption{Effect of multi-class handling (pedestrian vs. riders) on detection performance ($LAMR$) for the reasonable scenario. The ''enforce'' (''ignore'') settings involves (not) penalizing samples of the other class for being categorized as the respective class. The first row (baseline) involves a single class, the second and third row involve two classes.}
\label{tab:detail_classes}
\centering
\begin{tabular}{@{} l c c c c c @{}}
\toprule
& \multicolumn{4}{c}{Test} \\
\cmidrule{2-5}
& \multicolumn{2}{c}{pedestrians} & \multicolumn{2}{c}{riders}\\

               Training & ignore & enforce & ignore & enforce \\
\hline
Baseline (pedestrians)                         & 9.4 & 11.0 & -    & -    \\
+Riders only                      & \textbf{9.3} & \textbf{10.3} & \textbf{8.8}  & \textbf{10.8} \\
+Riders with ride-vehicle                    & 9.4 & 10.5 & 11.1 & 12.2 \\
\bottomrule
\end{tabular}
\end{table}

\paragraph{Accuracy}
Here we evaluate to what degree our annotation accuracy requirements from Section \ref{sub:dataset-annotation} were actually met in practice in the final EuroCity Persons annotations.

To estimate the amount of missed annotations, we compare these with the object detector output. At a $fppi$ of 0.3 for Faster R-CNN$_{all}$ on the reasonable setup we manually count 230 missed annotation larger than 32 \textit{px}. However, the miss-rate for Faster R-CNN$_{all}$ at this $fppi$ is about 10\% for the small test scenario and about 30\% for the occluded test scenario. Using the more conservative 30\% figure, we estimate that, in fact, there are additional 99 missed annotations for pedestrians larger than 32 px, bringing the total missed annotation to 329. As there are about 48000 pedestrians in the test dataset, this corresponds to 0.7\% missed annotations, which lies within the 1\% quality requirement of Section \ref{sub:dataset-annotation}.

\begin{figure}[t]
\begin{center}
   \includegraphics[width=0.98\linewidth]{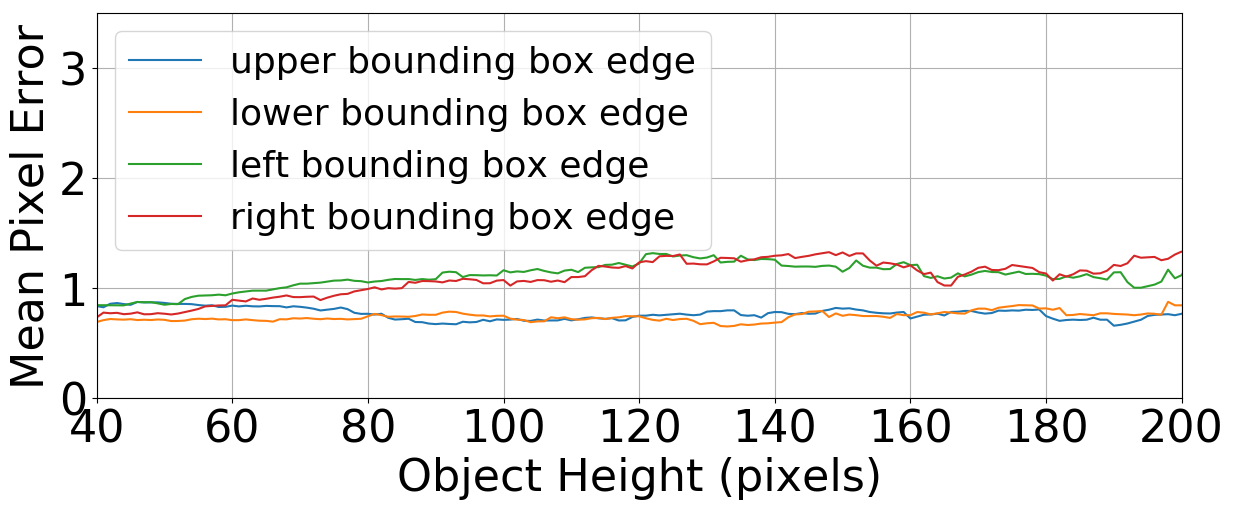}
\end{center}
   \caption{Mean pixel error between median of three additional annotators and the ECP dataset annotations, in dependence of object height p (averaged over the interval [p-20, p+20]).}
\label{fig:bboxannotation}
\end{figure}

To determine the inter-annotator agreement and thus obtain an indication about achieved accuracy with respect to bounding box localization and orientation annotation, a random subset of 1000 not occluded pedestrians was labeled again by three different persons. We analyze the average deviation between the median value of the three annotators and the corresponding Eurocity Persons annotation, in dependence of the object size. Figure \ref{fig:bboxannotation} shows that the average deviation of the bounding box extents stays below 1.4 \textit{px} for objects up to 200 \textit{px} high (interestingly, upper/lower box side more accurate than left/right side). Figure \ref{fig:OrientationEstimation} shows that in terms of orientation angle, the average deviation starts at 20 degrees for object sizes of 40 \textit{px} and reduces to about 10 degrees for object sizes larger than 100 \textit{px}. We note that this lies within the requirements of Section \ref{sub:dataset-annotation} as well.

We now artificially disturb the annotation quality of the training dataset in the following experiments, see Table \ref{tab:accuracy}.
First, we randomly delete bounding boxes of instances and groups to simulate the effect of missed objects during annotation (''delete''). Second, we move bounding boxes by four pixels up or down and left or right (''jitter''). Third, we add (erroneous) ground-truth boxes to simulate the effect of hallucinating objects during annotation (''hallucination''). For this, a selected ground-truth bounding box itself is not changed but an additional, identically sized bounding box of the pedestrian class is placed at a random location in the image. Lastly, we introduce hallucinations that are more likely to resemble pedestrians, by running a SSD model of an early training stage on the training dataset (after 80000 iterations). The 11000 highest scoring false positives of these detections (corresponds to 10\% of all pedestrians in the training dataset) are handled as regular groundtruth boxes and added to the training dataset for the ''false positives'' experiment.
We examine different levels of disturbances by manipulating different amounts of bounding boxes. The effects for disturbances that are even worse than in our very first pilot study are also evaluated. The probability for a bounding box to be disturbed is given in the Table \ref{tab:accuracy}.

The detection performance of Faster R-CNN suffers from deleting and disturbing the bounding box locations.
Deleting 25\% of the bounding boxes results in a miss-rate of 11.7.
Note that with 75\% of the training samples a $LAMR$ of 10.0 is achieved (see Figure \ref{fig:exp_quantity}).
Pedestrians without bounding box labels may be used as background samples during training which results in the confusion of pedestrians and background during testing.
This effect is even stronger when OHEM is applied as seen when comparing R-FCN results with and without OHEM.
Placing hallucinations at random locations only slightly influences the overall detection performance.
Adding 10\% hallucinations that more resemble pedestrians (''false positives'') result in a more significant drop in performance of 3.3 points.

\begin{table}[!htp]
\small
\vspace*{5pt}
\caption{Perturbation analysis of annotation, effects on performance.}
\label{tab:accuracy}
\centering
\begin{tabular}{@{} l c c c c  @{}}
\toprule
Method & Disturbance & Prob. & LAMR & $\Delta$ \\
\hline
Faster R-CNN & none & -           & 9.4 & - \\
Faster R-CNN & delete & 10\%      & 10.2 & +0.8 \\
Faster R-CNN & delete & 25\%      & 11.7 & +2.3 \\
Faster R-CNN & false positives & 10\% & 12.7 & +3.3 \\
Faster R-CNN & hallucination & 20\% & 9.4 & +0.0 \\
Faster R-CNN & hallucination & 50\% & 9.9 & +0.5 \\
Faster R-CNN & jitter & 10\%     & 9.6 & +0.2 \\
Faster R-CNN & jitter & 20\%     & 9.8 & +0.4 \\
Faster R-CNN & jitter & 50\%     & 12.3 & +2.9 \\
R-FCN OHEM   & none & -           & 12.1 & - \\
R-FCN OHEM   & delete & 25\%      & 15.1 & +3.0 \\
R-FCN NoOHEM & none & -           & 12.2 & -   \\
R-FCN NoOHEM & delete & 25\%      & 13.9 & +1.7 \\
\bottomrule
\end{tabular}
\end{table}
{\tiny }

\section{Discussion}
\label{sec:Discussion}

\begin{figure*}
\begin{center}
   \includegraphics[width=0.3\linewidth]{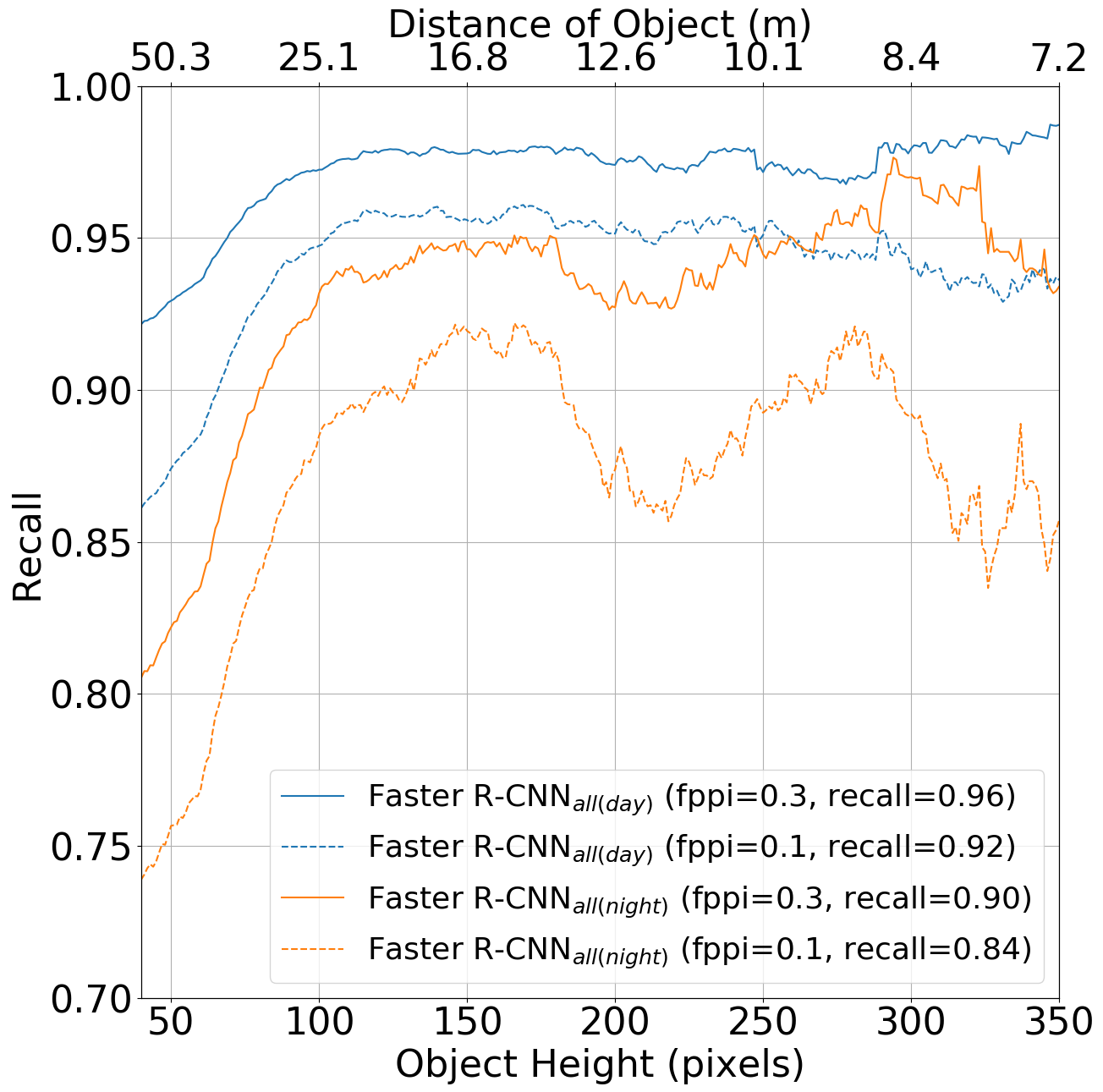}
	 \includegraphics[width=0.3\linewidth]{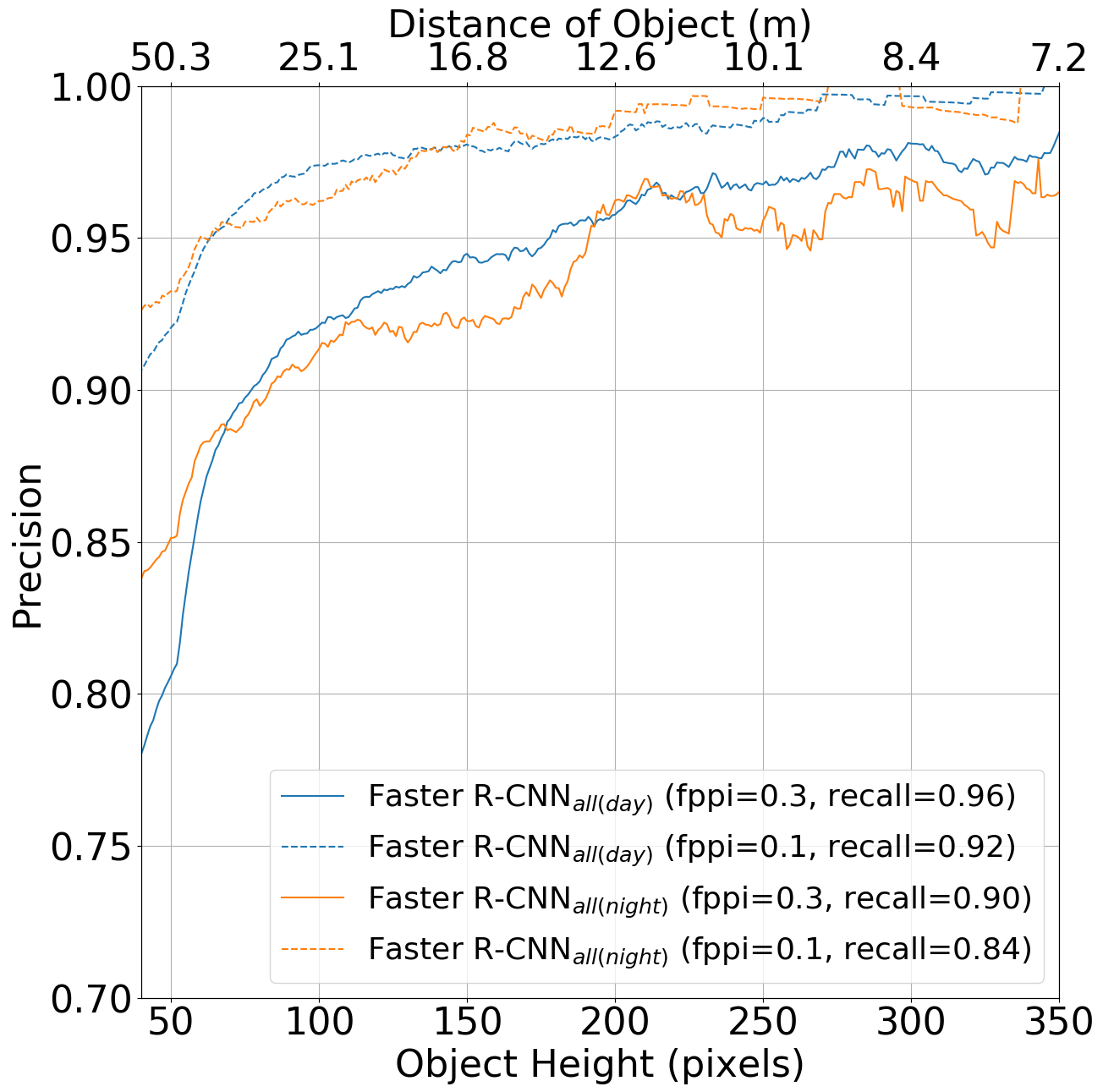}
		\includegraphics[width=0.3\linewidth]{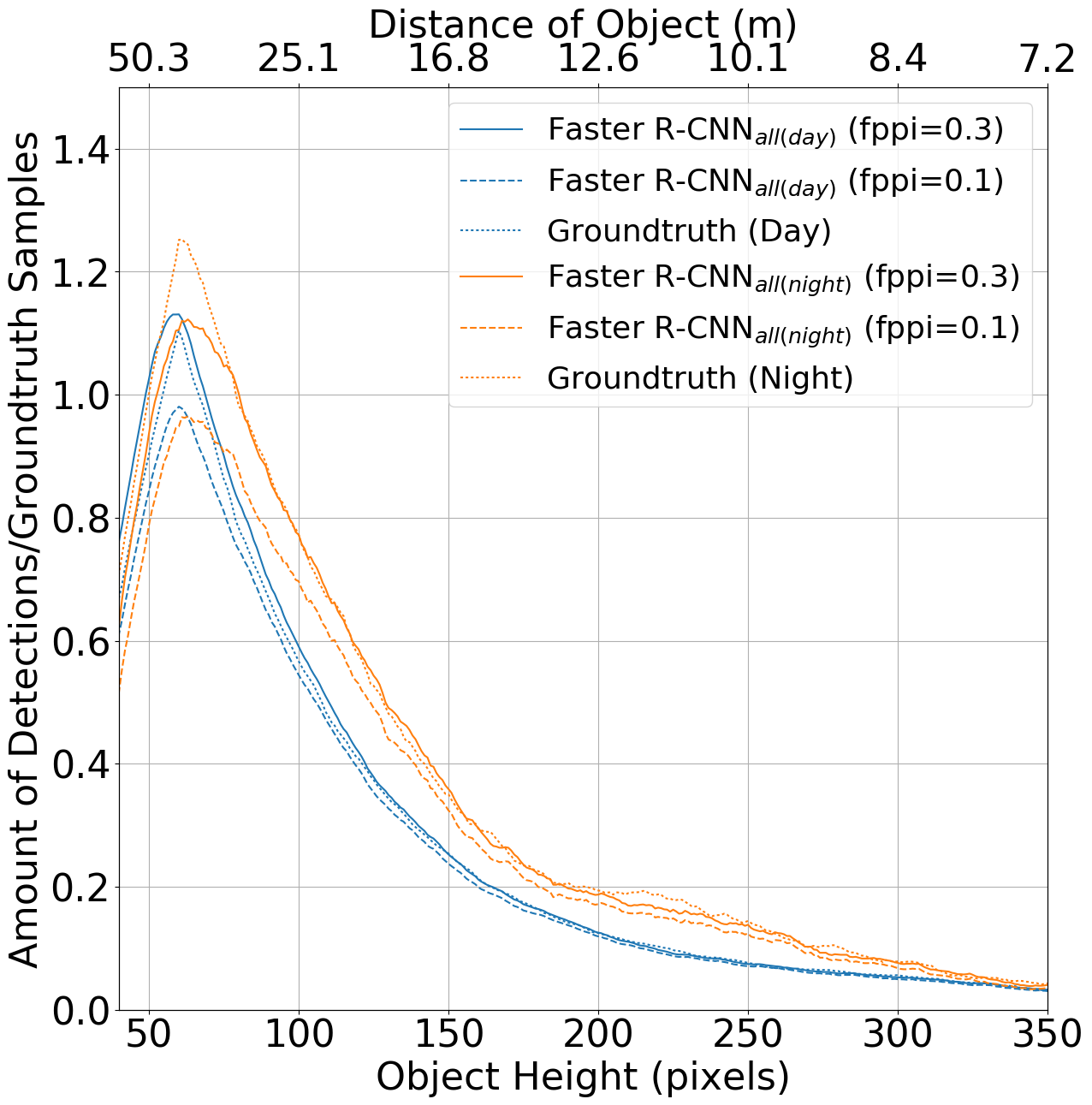}
\end{center}
   \caption{Recall (left), precision (middle) and the associated per-image detection and ground-truth sample counts (right) vs. object height at two operating points for the Faster R-CNN variant at day- and night-time (each trained and tested separately on upscaled day- and night-time images of EuroCity Persons reasonable). To calculate the distance of an object (upper x-axis) the camera calibration is used and a fixed object height of 1.7 m is assumed. For smoothing reasons, the recall and precision for object height p in pixels (\textit{px}) is computed within the height range [p-20 \textit{px}, p + 20 \textit{px}].}
\label{fig:relevant_pedestrians}
\end{figure*} 

A main outcome from the EuroCity Persons (ECP) experiments is that data still remains a driving factor for the person detection performance in urban traffic scenes: Even at training data sizes that are about one order of magnitude larger than existing ones (cf. Table \ref{tab:benchmarks_det}), the considered state-of-the-art deep learning methods (Faster R-CNN and SSD) do not saturate in detection performance.

The fact that saturation does not occur can be attributed to the diversity of the data. The ECP dataset covers a large geographical region, day and night, and different weather conditions. This quality is reflected in its generalization capability across datasets: pre-training on ECP and fine-tuning (post-training) on a smaller target dataset (KITTI, CP) yields better results than training solely on the target dataset. See Tables \ref{tab:results_pretrain_kitti} and \ref{tab:results_pretrain_cp}. Pre-training results in an increase of 5, 8, and 11 percentage points in average precision for the easy, moderate and hard KITTI validation datasets, respectively, when compared to using the original KITTI training data set. These benefits are larger than pre-training with CP. Similarly, pre-training results in a decrease of 2, 9, and 7 percentage points in $LAMR$ for the reasonable, small, occluded CP validation datasets, respectively, when compared to using the original CP training data set. Pre-training with ECP is especially valuable for the hard or occluded cases, involving improvements of about 10 percentage points in $LAMR$. 

On the other hand, existing biases in current databases are too strong to allow avoiding a fine-tuning step on the target dataset altogether;
this holds even for a much larger source training dataset like ECP. Current deep learning methods need to adapt to these biases first, even if the same vision task is involved. This is in accordance with earlier experiments on transfer learning on a smaller scale by \cite{ben2017CityPerson}.  

The ECP dataset allowed us to analyze some biases in more detail. Foremost, experiments suggest that there is indeed a bias derived from large geographical region. We compiled datasets for central West Europe vs. central East Europe, where other factors influencing performance were held similar. 
We found that the existence of a bias is statistically significant with a confidence interval of 99\%. 

Comparing day- and night-time detection performance, one observes from Table \ref{tab:nighttime_daytime} that at equal training set sizes, night-time performance is worse (a $LAMR$ of four points higher). This difference is enlarged when the entire day- and night-time training sets of ECP are used as the former is an order of magnitude larger. See Figure \ref{fig:relevant_pedestrians}. The drop in recall for pedestrians closer than 8 m could be due to the headlights of the recording vehicle. These could result in very bright spots for the lower body of pedestrians and complicate detection. Our dataset provides the possibilities to further research in this direction and compare differences between day and night recordings.
  
The way annotations are performed proves to be important as well. As in \cite{ben2017CityPerson} we show that a correct ignore region handling has an impact on detection performance. In our case it boosts performance by $1.5$ points (see Table \ref{tab:detail_ignore}). This is a larger difference than that between the performances using 75\% and 100\% of the training data in Figure \ref{fig:exp_quantity}. We go beyond \cite{ben2017CityPerson} to show that it is beneficial to train specific detectors for classes that otherwise might be confused with the target class. In our experiments, the jointly trained detection models for riders and pedestrians achieve a lower miss rate for the pedestrian class, than models trained for pedestrians-only, when the precise class is enforced. In the evaluation protocol of \cite{ben2017CityPerson} this case is not considered as riders are always handled as ignore regions.

It is interesting to put current person detection performance in context. When viewed in historic context, the best-performer on an early benchmark \cite{enzweiler2009monocular} was a method based on HOG features and SVM classifier. When comparing its performance with that of the best-performer in this paper, the R-CNN, one observes that performance has improved by an order of magnitude over the past decade, in terms of the reduction of the number of false positives at given correct detection rate, albeit dealing with two different datasets of urban traffic (Figure 8 in \cite{enzweiler2009monocular} vs. Figure \ref{fig:missrate_curves} here).

In this paper, we focused on generic person detection performance, but also the application context can be considered. State-of-the-art detection performance (e.g. correct detection around 90\% at $0.1 - 0.3~fppi$) is sometimes cited as evidence that performance is far away from practical use for the vehicle application. This is incorrect, as can be readily inferred from the fact that there are already several vision-based person detection systems on-board production vehicles on the market. A number of factors improve performance in the vehicle application. First, other than we assume in this study, not all errors are equal in the vehicle application. Errors increasingly matter when they involve objects close to the vehicle. The detectors improve their performance with decreasing distance (increasing object size). See Figure \ref{fig:relevant_pedestrians}, the detection rate increases to 97\% at a distance of 25 $m$ (object height 100 \textit{px}). Second, some false positives can be eliminated, when taking advantage of known scene geometry constraints (e.g. pedestrians or riders should be on the ground plane, their heights should be physically plausible when accounting for perspective mapping). Third, many false positives arise by an accidental overlaying of structures at different depths, and are not consistent over time when observed from a moving camera. Tracking can suppress such false positives (\cite{enzweiler2009monocular} shows a reduction by up to 37\%). Last but not least, active safety systems for pedestrians and cyclists involve additional sensors for detecting obstacles in front of the vehicle: a second camera (stereo vision), radar or LiDAR. Thus vehicle actuation (braking, steering) does not solely rely on monocular object detection. It should be finally noted that current commercial systems are in the context of driver assistance, meaning that a correct detection performance of about 90\% is acceptable, as long as the false alarm rate is essentially zero. 

This brings us to the human baseline. A visual inspection shows that the remaining errors are indeed ''hard'', even for a human, see Table \ref{tab:qualitative_mistakes}. A recent paper \cite{zhang2016tpami} finds that current single-frame pedestrian detection performance lags that of an attentive human by an order of magnitude. Thus there is a potential for a substantial further performance improvement; an improvement which would be important with the advent of fully self-driving vehicles. 

More data remains part of the solution on how to improve performance.
Our study shows that performance still improves with increasing training set size with a decent gradient (i.e. Figure \ref{fig:exp_quantity}). A further doubling of the current training size (110000 pedestrians) is projected to yield a reduction of the $LAMR$ from 9.4 to about 7.6 points. More training data is especially helpful for persons in non-standard poses, in rainy or night-time conditions, or under partial occlusions. The found relations between annotation quality and quantity on one hand and detection performance on the other (i.e. Table \ref{tab:accuracy}), together with a price tag for annotations at various quality levels can help optimizing the requirement specification for dataset annotation.

In terms of vision methods, better solutions are needed to provide accurate localization in the presence of multiple persons and significant occlusion. Recent detection methods like R-FCN or Faster R-CNN have profited from incorporating the proposal generation in an end-to-end learning strategy. Still, the proposal boxes are classified independently of each other resulting in multiple detections for the same object in particular if the proposals share similar image locations. In general, there is no loss enforcing a one to one matching between detections and ground-truth samples. The task of suppressing multiple detections for the same object is usually solved by the decoupled non-maximum suppression. Interestingly, most top performing methods of the common generic object detection benchmarks depend on a simple greedy non-maximum suppression (NMS) \cite{BodlaSCD17}. This NMS  poses a problem for overlapping objects e.g. in pedestrian groups.
When selecting the \textit{IoU} threshold there is a tradeoff between recall and precision as shown in Table \ref{tab:qualitative_mistakes}. In \cite{Hosang2017cvpr} a neural network is trained to rescore detections, which renders a further NMS stage unnecessary. Still, the neural network solely relies on bounding box locations and confidence scores as input. To further improve the performance \cite{Hosang2017cvpr} proposes to incorporate image features in future works. Doing so the network could be informed about how many objects are present. As it is already a neural network architecture it can easily be integrated into existing detection networks. Thus, the three steps proposal generation, classification/bounding box regression, and NMS would finally be combined in a true end-to-end approach.

A number of methodical avenues could improve classification performance. In Figure \ref{fig:relevant_pedestrians} and in our baseline experiments we show that  small objects are still very challenging despite the great amount of small sized pedestrians present in our training dataset. Approximately 75\% of the false positives at 0.3 $fppi$ analyzed in Figure \ref{fig:quantitative_failure_analysis} are smaller than 80 pixels. Recently, methods have been published that are tuned for the detection of smaller objects like MS-CNN. Such methods have to be analysed in detail to find still remaining weaknesses and further possibilities for improvement. We show quantitatively in Figure \ref{fig:quantitative_failure_analysis} and qualitatively in Table \ref{tab:qualitative_mistakes} that depictions, reflections and clothes are often confused with real pedestrians. These confusions result in high scoring false positives also for sizes larger than 80 \textit{px}. That necessitates the design of appropriate multi-task deep nets that more effectively incorporate global scene context. When training a detection network jointly for pedestrians and riders we have already shown that confusions between the two person classes can be reduced. Utilizing the already annotated reflections and depictions as additional classes during training could improve the discrimination performance as well.
An ensemble of specialized deep learning models could take advantage of known bias (particular location and digital maps, weather, time of day). Such an approach could even switch on a per frame basis between sub-models, e.g. when there is a sudden change in lighting. For example lenseflares might occur from one frame to another when the vehicle turns into the direction of the sun.

As person detection is being perfected, the focus of research will likely shift to tracking and motion prediction.  Motion prediction based on point kinematics is often not accurate because of abrupt changes in person motion. Systems like \cite{kooij2014context} come into play which take into account additional pose information. In preparation for this, we included in this benchmark the orientation estimation of the overall body, and showed that the latter can be jointly trained with the detection task at minimal performance loss.

\section{Conclusions}

We have created the new EuroCity Persons dataset, which takes annotations of persons in urban traffic scenes to a new level in terms of quantity, diversity and detail. 
We optimized four state-of-the-art deep learning approaches (Faster R-CNN, R-FCN, SSD and YOLOv3) to serve as baselines for the new person detection benchmark; we found a variant of Faster R-CNN to perform overall best, with a log-average-miss-rate of 8.1, 17.1 and 33.9 on the reasonable, small and occluded test scenario, respectively.  

The experiments show that data is still a driving factor for the person detection performance in urban traffic scenes: Even at the new training data sizes that are about one order of magnitude larger than previous ones, the considered deep learning methods do not saturate in detection performance. This can also be attributed to the diversity of the dataset. In experiments on transfer learning, we showed that 
detectors pre-trained with the new dataset and fine-tuned on a target dataset, yield superior performance than those trained on the target dataset only (improvements on KITTI and CityPersons by 5-11 and 2-9 points, respectively). 
 
The experiments also showed that night-performance is a few percentage points lower than day-time performance. Experimental results furthermore indicate that a statistically significant bias exists on detection performance across large-scale regions in Europe, resulting in performance variations of the same order. Adding orientation estimation to object detection lowers the detection performance by a single percentage point for the Biternion loss. 

System performance regarding person detection in urban traffic settings has improved by an order of magnitude over the past decade;
it is now closing in on human performance. Future improvement will in part still come from additional data. More person training data is especially helpful in non-standard poses, in rainy or night-time conditions, or under partial occlusions. We provided some insights regarding the effect of annotation accuracy on performance that could be useful for future annotation efforts. The development of appropriate multi-task deep networks, which combine a holistic approach to scene understanding with specialized person detection, taking advantage of known bias (geolocation, time of day, weather condition) seem promising. We hope that the new EuroCity Persons benchmark will stimulate research towards finding the ''perfect'' person detector.

\ifCLASSOPTIONcaptionsoff
  \newpage
\fi



\bibliographystyle{IEEEtran}
\bibliography{IEEEabrv,egbib}
%
%


%

\begin{IEEEbiography}[{\includegraphics[width=1in,height=1.25in,clip,keepaspectratio]{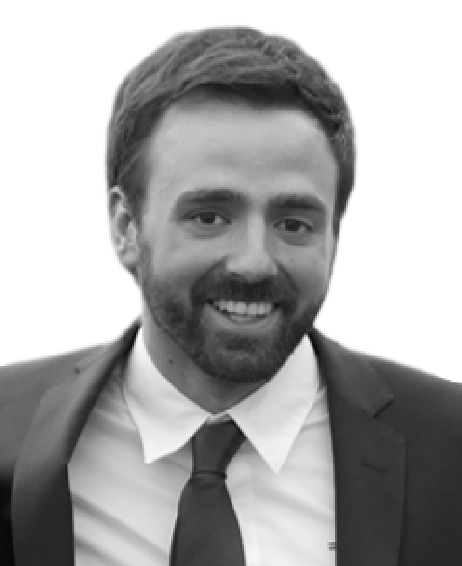}}]{Markus Braun}
Markus Braun received the M.Sc. degree in computer science from Karlsruhe Institute of Technology, Germany, in 2015. Since then he is working toward the Ph.D. degree at TU Delft, Delft, The Netherlands. He is also currently with Daimler Research and Development in the Environment Perception department, Ulm, Germany. His research interests include machine learning and video analysis for automated driving, with a focus on detection and pose estimation of vulnerable road users.
\end{IEEEbiography}

\begin{IEEEbiography}[{\includegraphics[width=1in,height=1.25in,clip,keepaspectratio]{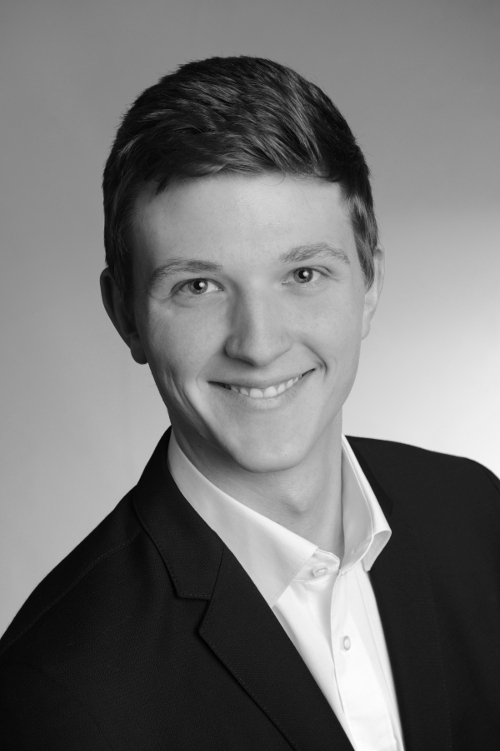}}]{Sebastian Krebs}
Sebastian Krebs received the M.Sc. degree in computer science from the University of Ulm, Germany, in 2016. Since then he is working toward the Ph.D. degree at TU Delft, Delft, The Netherlands. He is also currently with Daimler Research and Development in the Environment Perception department, Ulm, Germany. His research interests include machine learning and video analysis for automated driving, with a focus on vulnerable road user tracking.
\end{IEEEbiography}


\begin{IEEEbiography}[{\includegraphics[width=1in,height=1.25in,clip,keepaspectratio]{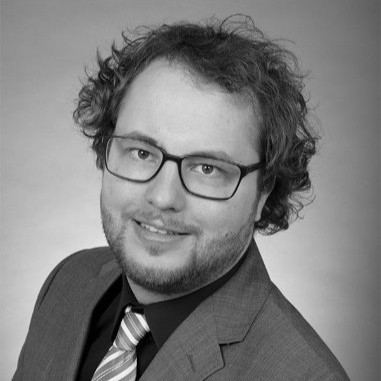}}]{Fabian Flohr}
received the M.Sc. degree in computer science from Karlsruhe Institute of Technology, Germany, in 2012. He is currently working toward the Ph.D. degree at the University of Amsterdam, The Netherlands. He is also currently with Daimler Research and Development, Ulm, Germany, where he has the technical lead for vulnerable road user sensing.
\end{IEEEbiography}

\begin{IEEEbiography}[{\includegraphics[width=1in,height=1.25in,clip,keepaspectratio]{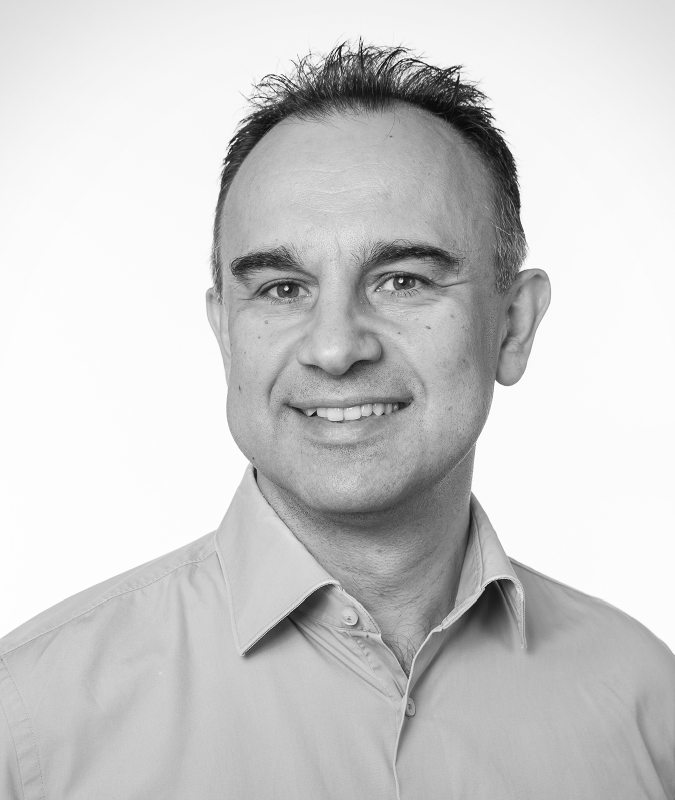}}]{Dariu M. Gavrila}
received the Ph.D. degree in computer science from Univ. of Maryland at College Park, USA, in 1996. From 1997 until 2016, he was with Daimler R\&D, Ulm, Germany, where he became a Distinguished Scientist. He led the multi-year pedestrian detection research effort at Daimler, which was incorporated in the Mercedes-Benz S-, E-, and C-Class models (2013-2014).  
He was awarded the Outstanding Application Award 2014 from the IEEE Intelligent Transportation Systems Society, as part of a Daimler team. 
In 2003, he also became a part-time Professor with Univ. of Amsterdam, in the area of intelligent perception systems. Over the past 20 years, he has focused on visual systems for detecting human presence and activity, with application to intelligent vehicles, smart surveillance, and social robotics. In 2016, he moved to TU Delft, where he since heads the Intelligent Vehicles group as a full-time Professor.
\end{IEEEbiography}




\end{document}